\documentclass[runningheads]{llncs}
\usepackage[T1]{fontenc}
\usepackage{natbib}
\usepackage{subfigure}
\usepackage{amsmath}
\usepackage{amssymb}
\usepackage{multirow}
\usepackage{graphicx}
\begin{document}
\title{KANFormer for Predicting Fill Probabilities via Survival Analysis in Limit Order Books}
%

\author{Jinfeng Zhong\inst{1,2}, 
Emmanuel Bacry\inst{2}, Agathe Guilloux\inst{3}, Jean-François Muzy\inst{4}}
\authorrunning{Zhong et al.}
%
\institute{LIASD – Université Paris 8, 93200 Saint-Denis \email{jinfeng.zhong@univ-paris8.fr}\and Ceremade, CNRS-UMR 7534, Université Paris-Dauphine PSL, Place du Maréchal de Lattre de Tassigny, 75016 Paris, France \and Inria, Université Paris Cité, Inserm, HeKA (UMR 1346), 75015 Paris, France \and Laboratoire ``Sciences Pour l'Environnement'' UMR 6134 CNRS - Université de Corse - Campus Grimaldi, 20250 Corte (France)}

\maketitle              
\begin{abstract}
This paper introduces KANFormer, {a novel deep-learning-based model for predicting} the time-to-fill of limit orders by leveraging both market- and agent-level information. KANFormer combines a Dilated Causal Convolutional network with a Transformer encoder, enhanced by Kolmogorov–Arnold Networks (KANs), which improve nonlinear approximation. Unlike existing models that rely solely on a series of snapshots of the limit order book, KANFormer integrates the actions of agents related to LOB dynamics and the position of the order in the queue to more effectively capture patterns related to execution likelihood. We evaluate the model using CAC 40 index futures data {with labeled orders}. The results show that KANFormer outperforms existing works in both calibration (Right-Censored Log-Likelihood, Integrated Brier Score) and discrimination (C-index, time-dependent AUC). We further analyze feature importance over time using    {SHAP (SHapley Additive exPlanations)}. Our results highlight the benefits of combining rich market signals with expressive neural architectures to achieve accurate and interpretabl {predictions} of fill probabilities.

\keywords{Limit order book;  Survival analysis; Fill probability prediction; Order submission; High-frequency trading.}
\end{abstract}

\section{Introduction}

Electronic financial exchanges commonly use limit order books (LOBs) to match buy and sell orders across various asset classes. Among the order types supported by LOBs, \textit{limit orders}, \textit{market orders} and \textit{cancel orders} are the most frequently used \citep{abergel2016limit, gould2013limit}. {Unlike market orders, which execute immediately at the prevailing bid or ask, limit orders are placed at predefined prices to seek a more favorable execution.} Although limit orders can yield more favorable prices, they carry the risk of non-execution: these orders remain in the book until they are either matched with an incoming market order or canceled before execution. Therefore, it is crucial for investors to assess the likelihood that a limit order will be executed within a given time horizon. The time a limit order remains in the book before execution, referred to as its time-to-fill, can vary widely and depends on the dynamics of the LOB \citep{arroyo2024deep,maglaras2022deep}. {Predicting the time-to-fill distribution or, equivalently, fill probabilities of limit orders} plays a critical role in optimal execution strategies and has attracted significant research attention \citep{cho2000probability, LO200231, maglaras2022deep, arroyo2024deep}.

The time-to-fill of a limit order is inherently subject to right-censoring \citep{lagakos1979general}: many orders are canceled or expire (i.e., remain in the LOB until the end of the trading day) without being executed, leaving their true fill times unobserved. This makes survival analysis, a statistical framework for censored time-to-event data, a natural tool for modeling and predicting the time-to-fill of limit orders \citep{LO200231, maglaras2022deep, arroyo2024deep}.

{Survival analysis has a long tradition in statistics \citep{cox1972regression} and, more recently, has been combined with deep learning to handle high-dimensional data and improve predictive power \citep{kvamme2019time, rindt2022survival,bleistein2024dynamical}. Its application to limit order books began with econometric models of execution risk based on survival functions \citep{LO200231}, and was later extended to capture execution dynamics under censoring \citep{maglaras2022deep, arroyo2024deep}. Building on this line of work, deep learning approaches specifically tailored to LOB data have emerged, including recurrent neural networks (RNNs) \citep{maglaras2022deep} and Transformer architectures \citep{arroyo2024deep}.} However, these methods still face limitations in input representation, evaluation protocols, and interpretability, which we revisit in Section~\ref{sec:related_work}.

In this paper, we propose a new deep-learning-based survival analysis model, KANFormer, to model the fill probabilities of limit orders. Our key contributions are as follows:

\begin{itemize}
\item \textbf{Rich input features beyond LOB snapshots:} Prior models rely {exclusively} on LOB state features such as prices and volumes at different levels \citep{arroyo2024deep,maglaras2022deep}. We extend this representation by incorporating agent-level behavioral data, namely order submissions, cancellations, and modifications, which provide additional insights into market dynamics that are not observable from LOB snapshots alone.

\item \textbf{Comprehensive evaluation metrics:} To rigorously evaluate model performance, we employ both \textit{calibration} and \textit{discrimination} metrics, addressing shortcomings in prior work that relied on incomplete metrics \citep{maglaras2022deep,arroyo2024deep}. Specifically, we assess calibration using the \textit{Right-Censored Log-Likelihood (RCLL)} \citep{balakrishnan2012left} and the \textit{time-dependent Brier score} \citep{gerds2006consistent}, and discrimination using the \textit{C-index} \citep{uno2011c} and \textit{time-dependent AUC} \citep{lambert2016summary}. This dual approach ensures that the predicted survival distributions are both well-calibrated and properly ranked.

\item \textbf{KAN-enhanced Transformer architecture:} We propose a dual-encoder Transformer in which the standard feed-forward networks (FFNs) are replaced by \textit{Kolmogorov–Arnold Networks (KANs)} \citep{liu2024kan}. Inspired by recent applications of KANs to time series forecasting \citep{genet2024temporal, genet2024tkan, han2024kan4tsf, zhang2024transformer}, this design strengthens nonlinear approximation while preserving the attention backbone \citep{vaswani2017attention}.

\item \textbf{Model interpretability:} We employ    {SHAP (SHapley Additive exPlanations)} \citep{lundberg2017unified} to visualize global feature importance and track how different input features influence fill probabilities over time. This provides insight into the mechanisms driving execution risk and highlights the roles of both market and agent-level signals.
\end{itemize}

The remainder of this paper is organized as follows. 
Section~\ref{sec:related_work} reviews related literature and discusses the limitations of prior approaches. 
Section~\ref{sec:problem_formulation} formulates the prediction task as a survival analysis problem. Section~\ref{sec:model} introduces the KANFormer architecture. Section~\ref{sec:experiments} presents empirical results, including benchmark comparisons, 
ablation studies, and feature attribution analysis. 
Finally, Section~\ref{sec:conclusions_perspectives} concludes with future research directions.

\section{Limit Order Books and Survival Analysis}\label{sec:related_work}

This section provides the necessary background on limit order books (LOBs), 
survival analysis, and their application to modeling order execution. 
We begin with the operational structure of LOBs, then introduce the survival analysis framework 
and its relevance for execution modeling. 
We next review prior work and highlight their main limitations, which directly motivate our proposed approach.

\subsection{Limit Order Books}\label{sec:LOBRE}

LOBs are the primary mechanism by which modern exchanges facilitate trading across asset classes. 
A LOB displays all incoming orders with their price and quantity, matching buy and sell orders based on price–time priority. 
The bid side contains buy limit orders sorted by descending price, while the ask side contains sell orders sorted by ascending price. 
The best bid and ask define the top of the book, and their difference is the bid--ask spread \citep{bouchaud2009markets}.

Order matching typically follows a price--time priority rule: better-priced orders execute first, and among equal prices, earlier orders take precedence. 
Unexecuted limit orders enter a queue and wait to be matched or canceled. 
The time-to-fill of a limit order therefore depends on queue position, incoming order flow, and broader market activity \citep{cartea2015algorithmic}. 
A line of \textit{queue-reactive models} emphasizes this mechanism explicitly: 
\citet{huang2015simulating, wu2019queue} simulate order-level behavior by tracking positions in the queue, 
while \citet{cartea2014buy} highlight the importance of queue priority for optimal execution strategies. 
These studies underline queue position as a key determinant of fill probabilities, a feature we explicitly incorporate in our framework.

Because many orders are canceled or expire without execution, right-censoring is pervasive. This naturally motivates the use of survival analysis as a principled tool for modeling time-to-fill \citep{arroyo2024deep, maglaras2022deep}.

\subsection{Survival Analysis}\label{sec:survival_analysis_sota}

A common line of work casts LOB prediction as a classification problem at a fixed horizon $H$ (e.g., mid-price up/down/unchanged \citep{zhang2019deeplob}). While intuitive, this framing has important drawbacks for execution modeling: (i) it imposes an arbitrary horizon $H$, (ii) censored orders (canceled or still active at $H$) are discarded or misclassified, and (iii) it ignores the timing of execution beyond $H$.  
  
In contrast, survival analysis, a framework explicitly designed for time-to-event data under censoring \citep{kalbfleisch2002statistical}, directly models the time-to-fill distribution under censoring, providing both horizon-specific discrimination (e.g., time-dependent AUC \citep{kamarudin2017time}) and calibration of the full predictive distribution (e.g., Negative Right-Censored Log-likeLihood (RCLL)). {Formally, let us denote $T \in \mathbb{R}^{+}$ the duration between the upcoming execution time and the current time, with covariates $\mathbf{x} \in \mathbb{R}^p$.} Since $T$ is not always be observed, we introduce a censoring time $C \in \mathbb{R}^{+}$, which corresponds to cancellation or expiration at market close. The observed time is then
\[
T^C = \min\{T, C\},
\]
with censoring indicator
\[
\delta = \mathbf{1}\{T \le C\},
\]
where $\delta=1$ if the order is executed (partially or fully) and $\delta=0$ if it is censored (the order is canceled or expires before execution). 
A dataset of $n$ orders is thus represented as
\[
\{(\mathbf{x}_i, T^C_i, \delta_i)\}_{i=1}^n.
\]

It is central to predict the \emph{survival function}:
\begin{equation}
    S(t \mid \mathbf{x}) = \mathbb{P}(T > t \mid \mathbf{x}),
    \label{eq:survival_function}
\end{equation}
which gives the probability that an order remains unexecuted beyond duration $t$, {with $t$ being the duration measured relatively to the current time.}

In survival analysis, it is standard to assume that, conditional on covariates $\mathbf{x}$, the censoring time $C$ and the event time $T$ are independent \citep{leung1997censoring}. This assumption, known as \textit{independent censoring}, ensures that censored observations do not bias inference on execution times. Under this assumption, parameters can be optimized by minimizing the 
\emph{RCLL} \citep{kalbfleisch2002statistical}:
\begin{equation}
    \mathcal{L} = -\sum_{i=1}^n \Big[ \delta_i \log f(T^C_i \mid \mathbf{x}_i) + (1 - \delta_i) \log S(T^C_i \mid \mathbf{x}_i) \Big],
    \label{eq:log_likelihood}
\end{equation}
which properly accounts for both executed ($\delta_i=1$) and censored ($\delta_i=0$) orders 
\citep{arroyo2024deep}.

\subsection{Previous Works}\label{sec:survival_analysis}

The time-to-fill of a limit order has long been studied under {survival frameworks}.
\citet{handa1996limit} first modeled how traders balance price improvement against execution risk, and \citet{lo2002econometric} later applied survival analysis to LOB snapshots. More recently, deep learning models have been proposed to predict survival curves directly from LOB snapshots \citep{arroyo2024deep, maglaras2022deep}. For instance, \citet{arroyo2024deep} introduced a convolutional--Transformer encoder combined with a monotonic neural network decoder \citep{rindt2022survival}.

Different approaches have been adopted for evaluation. \citet{arroyo2024deep} utilized the RCLL \citep{kalbfleisch2002statistical}, which evaluates the overall quality of the predicted survival distribution. In contrast, \citet{maglaras2022deep} focused on discrimination metrics such as the time-dependent AUC \citep{hung2010estimation}, which measures the ability to distinguish between executed and unexecuted orders at a specific time horizon. Other metrics commonly employed include the Brier score \citep{graf1999assessment}, a horizon-specific measure of squared error, and the C-index \citep{harrell1996multivariable}, a global measure of concordance for order ranking. Former works typically adopt only a subset of these metrics: for example, some focus exclusively on time-dependent AUC or C-index, while others rely solely on RCLL.

Beyond evaluation metrics, recent advances have also focused on enhancing predictive performance through more sophisticated model design. Kolmogorov--Arnold Networks (KANs) \citep{liu2024kan}, originally introduced as spline-based universal function approximators, have shown strong performance in sequential prediction tasks \citep{genet2024temporal, genet2024tkan, han2024kan4tsf, zhang2024transformer}. Replacing the feedforward layers of Transformers with KAN blocks yields KAN-Transformers, which offer enhanced expressive capacity and generalization. To our knowledge, these architectures have not yet been applied to survival analysis or execution modeling.

{As far as interpretability is concerned, prior work has mainly relied on global feature importance techniques. 
In particular, \citet{arroyo2024deep} applied SHAP values to identify which LOB features contribute most to the predicted execution risk. 
Their analysis focuses on feature importance computed at observed event times, providing a static view of explanatory factors within the survival modeling framework.
}


\subsection{Limitations of Previous Works}\label{sec:limitations}

\begin{table}[t]
\centering

\begin{tabular}{|l|l|c|c|l|}
\hline
Metric & Type & Horizon-specific & Scope & Practical readability \\
\hline
RCLL & Calibration & No & Full distribution & Low \\
BS  & Calibration & Yes & Time-dependent & High \\
C-index & Discrimination & No & Global ranking & High \\
AUC  & Discrimination & Yes & Time-dependent & High \\
\hline
\end{tabular}
\caption{Comparison of survival model evaluation metrics.
RCLL evaluates calibration of the full predictive distribution,
whereas Brier score and time-dependent AUC assess performance at specific horizons.
The C-index is a global ranking measure.
{Practical readability refers to how easily the absolute value of a metric 
can be interpreted. For example, AUC and C-index correspond to probabilities 
of correct ranking, whereas the RCLL depends on data scale and therefore lacks 
a direct intuitive meaning in absolute terms.}}
\label{tab:comparison_metrics}
\end{table}

\begin{figure}[t]
\centering

\begin{minipage}{0.48\textwidth}
    \centering
    \includegraphics[width=\linewidth]{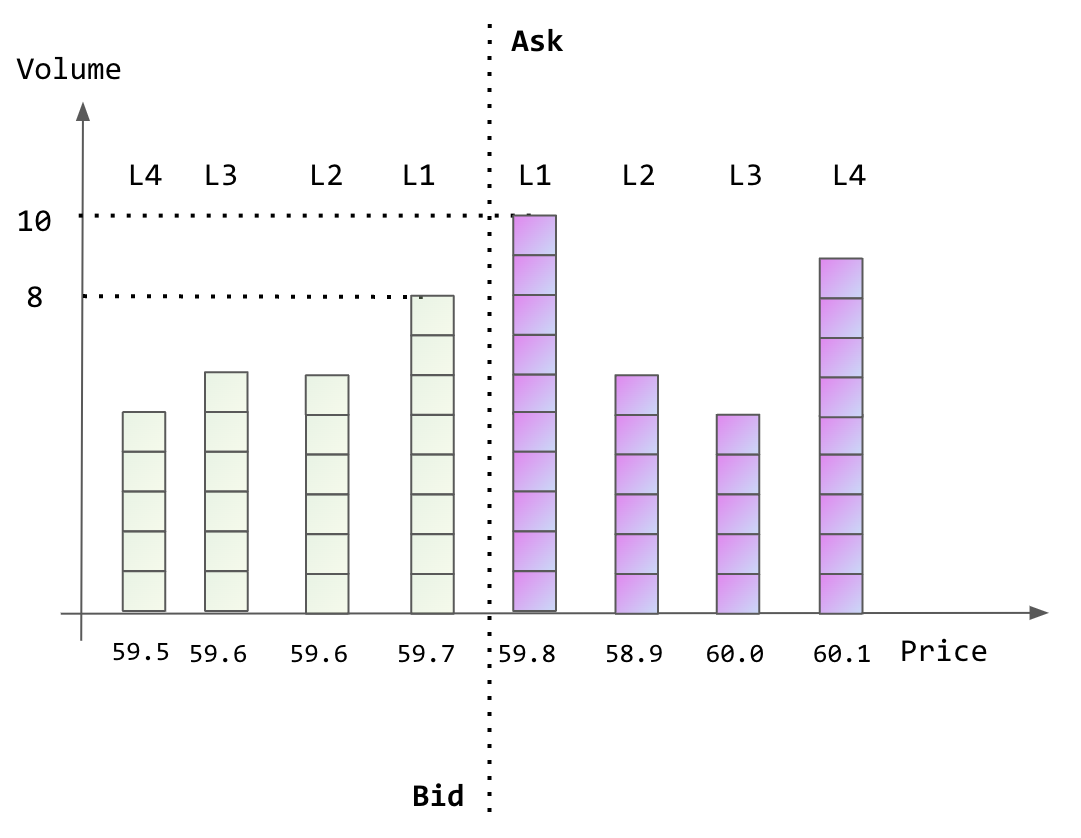}
    {\small (a) A snapshot of a LOB at $t$}
\end{minipage}
\hfill
\begin{minipage}{0.48\textwidth}
    \centering
    \includegraphics[width=\linewidth]{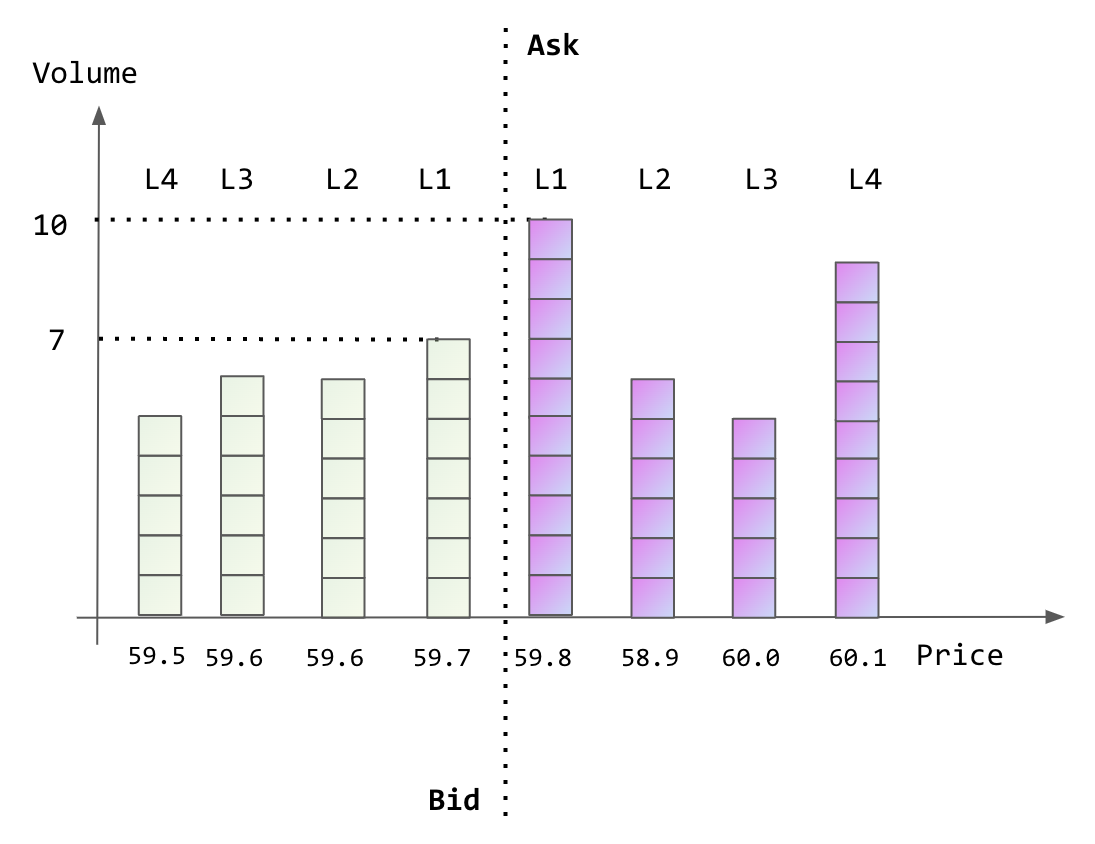}
    {\small (b) A snapshot of a LOB at $t+1$}
\end{minipage}

\caption{Two consecutive snapshots of a LOB. Each square represents a unit order; 
the purple bars denote ask (sell) orders, and the light green bars represent bid (buy) orders.}
\label{fig:lob}
\end{figure}

The discussions above highlight important foundations but also reveal limitations:
\begin{itemize}
    \item \textbf{Restricted representation of input:} Most existing models rely on LOB snapshots (e.g., price and volume levels) {as input data}, while neglecting agent-level actions. Figure~\ref{fig:lob} illustrates two consecutive LOB snapshots in which the volume at the best bid decreases from 8 to 7 between $t$ and $t + 1$. This change may arise from two distinct causes: (i) a market order consuming one unit at the best bid, or (ii) the cancellation of a standing bid order. Although the observed LOB update is identical, the underlying mechanisms, and their informational content, differ fundamentally.
    
    \item \textbf{Incomplete and hard-to-interpret evaluation protocols:} Time-dependent AUC, time-dependent Brier score, and C-index, although informative, may not capture the full quality of survival predictions \citep{rindt2022survival}. Relying solely on discrimination metrics such as the AUC or C-index can obscure calibration performance, i.e., whether predicted survival probabilities align with observed outcomes. Conversely, while some works emphasize the RCLL for its statistical rigor in evaluating the overall quality of the predicted distribution, its {practical readability} is low, making it difficult to interpret and compare in practice.
    \begin{quote}
        We use the term \emph{practical readability} to denote how easily a metric’s numerical value can be understood and compared in practice. For example, the Brier score corresponds to a mean squared error at a given horizon, and both the C-index and AUC can be interpreted as probabilities of correct ranking: concepts that are intuitive to practitioners. By contrast, the negative right-censored log-likelihood (RCLL) evaluates the fit of the entire predicted survival distribution. Although RCLL is a statistically principled \citep{rindt2022survival}, its absolute values depend on data scale and context, making them less directly comparable and harder to interpret in practice.
    \end{quote}
    Table~\ref{tab:comparison_metrics} summarizes the main properties of these metrics. Such incomplete evaluation protocols may lead to misleading conclusions. To address this, we adopt a comprehensive evaluation strategy that considers both calibration and discrimination. We provide further discussion on this issue, as well as the definitions of these metrics, in Appendix~\ref{appendix:metrics}.
    
    \item \textbf{Limited model expressiveness:} Transformer-based architectures have been used, but their reliance on standard feedforward layers overlooks newer approaches to modeling nonlinearity. More powerful alternatives, such as Kolmogorov–Arnold Networks (KANs) \citep{liu2024kan}, which offer enhanced expressive capacity and generalization \citep{genet2024temporal, genet2024tkan, han2024kan4tsf, zhang2024transformer}, remain unexplored in this context.
    
    \item {\textbf{Static interpretability:} Although prior work, such as \citet{arroyo2024deep}, applied SHAP to obtain global feature importance, their analysis was restricted to observed event times. This provides only a static snapshot of explanations and does not capture how the contribution of features evolves across prediction horizons, which is essential for dynamic trading strategies.}

\end{itemize}

These limitations motivate the need for a model that combines richer input features, more expressive architectures, comprehensive evaluation metrics, and dynamic interpretability. In Section~\ref{sec:model}, we address these challenges by integrating agent-level information, 
introducing \emph{KANFormer} that provides both accurate predictions and interpretable insights.

\section{{Setting the framework of modeling}}\label{sec:problem_formulation}

In this section, we formulate the estimation of fill probabilities as a survival analysis problem, 
following the framework introduced in Section~\ref{sec:survival_analysis_sota}. 
For each studied order at the best ask/bid level, the objective is to predict its survival curve, 
which gives the probability that the order remains unexecuted beyond any time horizon. 
Execution is defined as the order being at least partially filled, while censoring occurs 
if the order is canceled or expires before execution.

As introduced in Section~\ref{sec:survival_analysis_sota}, survival analysis provides a principled framework 
for modeling time-to-event data under censoring. In our setting, the event of interest is the first execution of a limit order, which may be partial or full. To formalize, for an order $i$ {placed on the current}, we denote (all times are expressed in seconds relative to midnight of the current day ):
\begin{itemize}
    \item \(t_i^S\): the submission time of the studied order $i$
    \item \(t_i^{0}\): the current time from which survival curve is modeled,
    \item $\delta_i$: label for execution ($\delta_i = 1$ if the order is executed on the current, otherwise $\delta_i = 0$)
    \item \(t_i^*\): time of cancellation (resp. execution) if the order is canceled (resp. executed)
    \item \(t_e\): time when market closes on the current
    \item \(T^C_i\): duration defined as  $T^C_i = t_i^* - t_i^{0}$ when the order is executed or cancelled, otherwise $T^C_i = t_e - t_i^{0}$
    
\end{itemize}

Formally, given the market signals $\mathbf{X}$ just before \(t_i^{0}\), the goal is to learn a model that outputs the survival function for all  \(t>t_i^{0}\).
\[
S(t \mid \mathbf{X}) = \mathbb{P}(T > t \mid \mathbf{X}).
\]
This formulation follows directly from the general setup in Section~\ref{sec:survival_analysis_sota}, 
but specializes it to the microstructural context of LOBs.

\section{The KANFormer model}\label{sec:model}

Building on the framework of modeling presented in 
Section~\ref{sec:problem_formulation}, we now introduce \emph{KANFormer}, 
our architecture for predicting the time-to-fill of limit orders. {We first present the features construction, then we introduce the structure of the KANFormer model.}

\subsection{Features construction}\label{sec:data_generation}

We describe how the {training} data is split and represented. The construction of each input sample relies on a series of historical LOB event streams. 
Specifically, we combine three blocks of features:  
(i) a series of actions submitted by market participants, which drive the evolution of the book;  
(ii) the corresponding LOB snapshots recorded at the exact moments when these actions are submitted, 
capturing the state of supply and demand; and  
(iii) the queue position of the studied order, which reflects its execution priority under 
price–time matching.  The dataset is generated from historical LOB event streams, 
which provide full information on all order submissions, cancellations, modifications, 
and executions.

Our procedure differs from the setting of \citet{arroyo2024deep}, 
where the survival function of a limit order is predicted from the instant of its submission 
at the best bid or ask. In contrast, we generalize the problem: 
we aim to predict the    {time-to-fill} of a limit order at the best bid/ask level 
from \emph{any subsequent moment} {($t_i^{0} \geq t_i^S$)}, not only from its submission time {($t_i^{0}= t_i^S$)}. 
A second difference is that we explicitly account for partial executions. 
In our dataset, more than 10\% of orders are partially executed before being 
canceled or expiring, which is not negligible. The generation process for each trading day is as follows:
\begin{enumerate}
    \item We randomly select orders submitted to the best bid or best ask level {of the LOB}. 
    \item We then randomly draw a time \(t_i^{0} \in [t_i^S, t_i^*)\). 
    If at \(t_i^{0}\) the order is still resting at the best level {of the LOB}, 
    we retain it as a valid sample; otherwise, we repeat the draw. 
    \item If, after five trials, the order is never observed at the best level {of the LOB} at the chosen \(t_i^{0}\), 
    the order is discarded. 
    \item For each retained order, we model its survival curve from time \(t_i^{0}\), 
    conditioned on the recent dynamics of the LOB and the order’s queue position. 
\end{enumerate}

Following \citet{arroyo2024deep}, we ensure that 100 valid orders are kept per trading day. The resulting dataset provides a rich set of input features from both the LOB and agent actions, 
as described below.

\paragraph{Features definition}  
The input data for our model is derived directly from historical LOB event streams 
and consists of three key sources of information:

     \textbf{Agent actions series:} A series of event messages submitted by agents, corresponding to the actions that generate the LOB updates before \(t_i^{0}\). 
    Each action is described by its type (e.g, insertion, cancellation, modification, market order, details provided in Appendix~\ref{appendix:Predictive_features}) as well as the 
    \emph{behavioral statistics of the submitting agent}. 
    Using the CAC 40 dataset from Euronext {that brings labeled orders}, we extract five representative measures for each agent over 300 trading days\textsuperscript{1}: Limit Ratio, Market Ratio, Cancel Ratio, Trade Ratio, and Aggressive Trade Ratio (see Appendix~\ref{appendix:Predictive_features} for details). These features provide rich contextual signals about the behavior of market participants 
    that are not visible from the LOB snapshots alone, see Section~\ref{sec:limitations}. Formally, the series of agent actions over a lookback window of length \( L \) is represented as
\[
\mathbf{A}_{\text{actions}} = \begin{bmatrix}
\mathbf{a}_{\text{actions}}(1) \\
\mathbf{a}_{\text{actions}}(2) \\
\vdots \\
\mathbf{a}_{\text{actions}}(L)
\end{bmatrix} \in \mathbb{R}^{L \times (d + 5)},
\]
where each row includes the action type embedding in $\mathbb{R}^d$
together with the five behavioral statistics of the submitting agent. 
    
    \textbf{LOB snapshot series:} For each action, we record the corresponding snapshot of the LOB at the moment the action is submitted. Each snapshot includes the top five price levels, 
    cumulative volumes on the same and opposite sides\textsuperscript{2}, 
    as well as derived statistics such as volatility, spread, volume imbalance, and time of day. 
    In total, 24 features describe each LOB snapshot; details are provided in 
    Appendix~\ref{appendix:Predictive_features}. The corresponding series of LOB snapshots is represented as
\[
\mathbf{X}_{\text{LOB}} = \begin{bmatrix} 
\mathbf{x}_{\text{LOB}}(1) \\
\mathbf{x}_{\text{LOB}}(2) \\
\vdots \\
\mathbf{x}_{\text{LOB}}(L)
\end{bmatrix} \in \mathbb{R}^{L \times (4n + 4)},
\]
where \( n = 5 \) price levels are considered, and the additional 4 columns represent 
the derived summary statistics.
    
    \textbf{Queue position:} The queue position of the studied order at \(t_i^{0}\). While LOB snapshots series and agent actions series capture the market dynamics before \(t_i^{0}\), queue position captures the studied order’s execution priority \citep{huang2015simulating, cartea2014buy} at \(t_i^{0}\). Each order is associated with a queue position, 
    denoted by $q$. For example, if four units are ahead of the studied order, then $queue = 4$.

\footnotetext[1]{Note that these agent-level statistics are not available to individual market participants. Therefore, a single agent cannot use the full model discussed in this work.}
\footnotetext[2]{Following \citep{maglaras2022deep}, we treat the market symmetrically. ``Same side'' is the side of the market to which the order is submitted; the ``opposite side'' is the other side.}

 In this paper, the length of the lookback window is set to $L=50$. {This choice is motivated by \citet{arroyo2024deep}, who compared \(L=50, 500,\) and \(1000\) and observed that larger windows do not necessarily improve calibration (RCLL) but substantially increase model size and training time. Hence, we adopt \(L=50\) as a practical compromise that captures local market dynamics while keeping computational cost manageable; this is also the length used by \citet{maglaras2022deep}.} Given \((\mathbf{A}_{\text{actions}}, \mathbf{X}_{\text{LOB}}, queue)\), we learn \(S(t \mid \cdot)\) from \(t_i^{0}\). KANFormer jointly encodes agent actions series, synchronized LOB snapshots series, and queue position to capture market dynamics and order priority:
\[
S(t \mid \mathbf{A}_{\text{actions}}, \mathbf{X}_{\text{LOB}}, queue) 
= \mathbb{P}(T > t \mid \mathbf{A}_{\text{actions}}, \mathbf{X}_{\text{LOB}}, queue),
\]
where \(T\) denotes the {time-to-fill} of the order. By jointly encoding these three inputs, KANFormer captures the market dynamics, the drivers of market dynamics, and the order’s microstructural priority.

In summary, unlike previous works \citep{arroyo2024deep, maglaras2022deep} that rely solely on a series of LOB snapshots, which may fail to capture the true dynamics of the market as discussed in Section~\ref{sec:limitations}, our representation integrates three complementary components. First, agent-level signals provide contextual information that helps recover the underlying causes of LOB changes. Second, LOB snapshots capture the state of supply and demand at the moments when these actions are submitted. Finally, queue position reflects the studied order’s execution priority under price–time matching \citep{huang2015simulating, cartea2014buy}. These three components are summarized schematically in Figure~\ref{fig:model}, which illustrates how they are later processed within the KANFormer presented in the next subsection.

\subsection{Model architecture}\label{sec:model_architecture}

\begin{figure}[t]
    \centering                                    
    \includegraphics[width=0.9\textwidth]{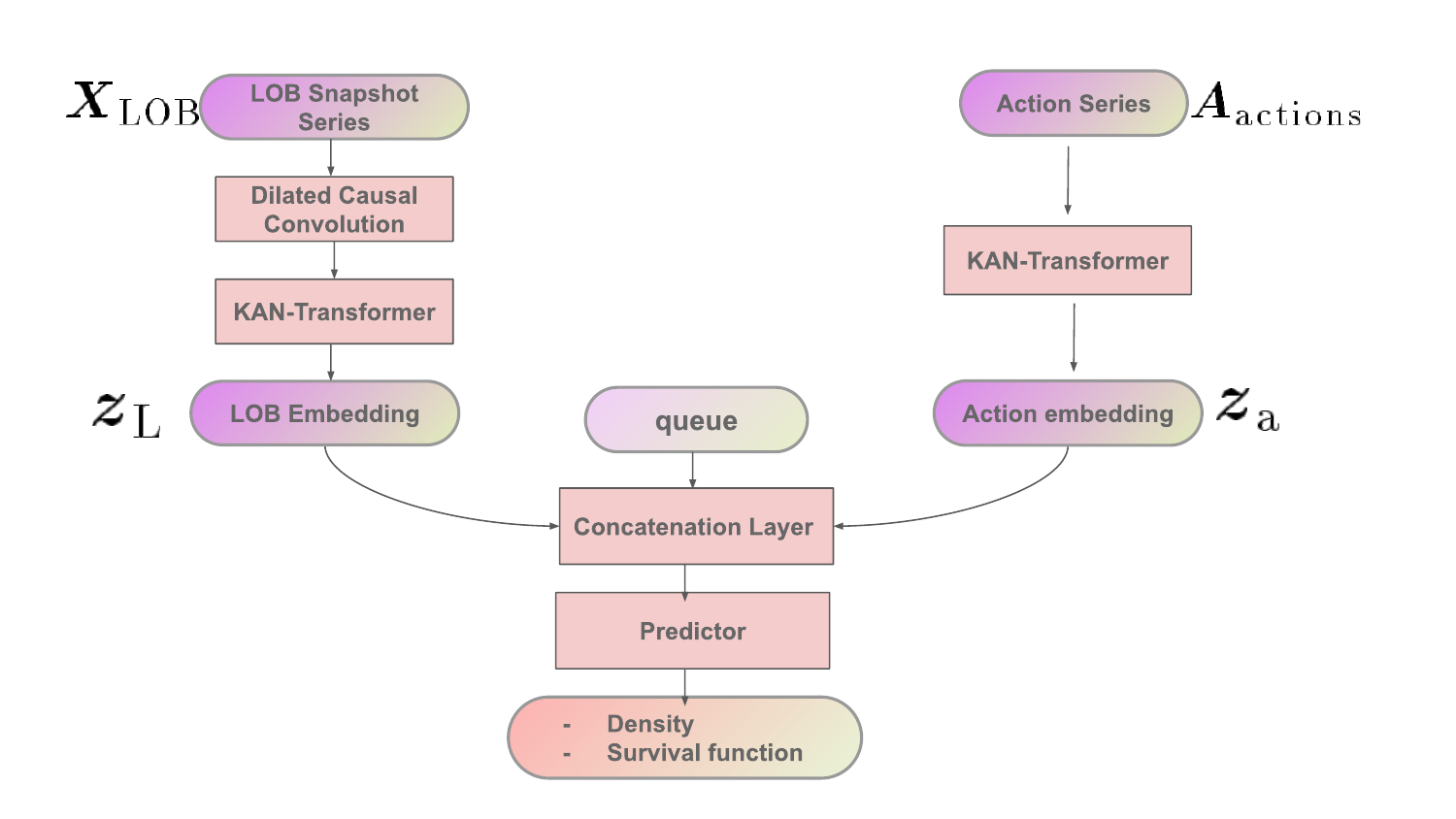}
    \caption{KANFormer architecture with Dilated Causal Convolution (DCC). LOB snapshots \(\mathbf{X}_{\text{LOB}}\in\mathbb{R}^{L\times(4n+4)}\) are processed by a DCC block followed by a KAN–Transformer to produce a LOB embedding. Agent actions \(\mathbf{A}_{\text{actions}}\in\mathbb{R}^{L\times(d+5)}\) are encoded by a KAN–Transformer to produce an action embedding. 
The two embeddings are {concatenated} with the queue position \(q\) and passed to the predictor, which outputs density and survival function.} 
    \label{fig:model} 
\end{figure}


Figure~\ref{fig:model} illustrates the overall structure of KANFormer. The model processes the LOB features and agent action sequences using two specialized 
KAN-Transformer encoders: the {Action KAN-Transformer} and the {LOB KAN-Transformer}. {Following \citet{arroyo2024deep}, we incorporate a Dilated Causal Convolutional (DCC) network \citep{van2016wavenet} within the LOB feature extractor. Specifically, DCCs process the LOB snapshots to generate queries, keys, and values, which are then passed to the Transformer self-attention layer. This locally-aware convolutional step provides contextualized features, enabling the Transformer to capture both short-term and long-range dependencies in the LOB time series.} The outputs of the {Action KAN-Transformer} and the {LOB KAN-Transformer} are then combined with the queue position of the studied order and passed to 
a {Predictor}, which produces the survival probability and density prediction.

{In principle, one could concatenate all features into a single input matrix 
\(\mathbb{R}^{L \times N}\), with \(N = d+5+4n+4\), and process them through a single KANFormer encoder. 
However, we adopt two specialized encoders, namely an {Action KAN-Transformer} and a {LOB KAN-Transformer}, to better reflect the heterogeneous nature of the inputs.
Agent actions (categorical embeddings and behavioral statistics) and LOB snapshots (price–volume and derived features) exhibit distinct temporal dynamics and statistical properties. Dedicated encoders allow the model to capture these patterns more effectively before combining them with the queue position in the predictor.} {We now introduce each module of the KANFormer model} {in Figure~\ref{fig:model}, the left part corresponds to the LOB KAN-Transformer that processes LOB snapshots series, the right part is the Action KAN-Transformer that processes the action series and the bottom part is the predictor that returns the survival function.}

\subsubsection{LOB KAN-Transformer}

The LOB feature matrix \( \mathbf{X}_{\text{LOB}} \in \mathbb{R}^{L \times (4n + 4)} \) 
represents the snapshot of the LOB observed at the exact moments when the actions are submitted. Each snapshot includes the top five price levels, cumulative volumes on the same and opposite sides, 
and derived statistics such as volatility, spread, volume imbalance, and time of day 
(see Appendix~\ref{appendix:Predictive_features}).  
The LOB KAN-Transformer processes this temporal sequence of snapshots to generate LOB embeddings:
\[
\mathbf{z}_{\text{L}} = \text{KAN-Transformer}(\mathbf{X}_{\text{LOB}}).
\]

\subsubsection{Action KAN-Transformer}

The agent action matrix \( \mathbf{A}_{\text{actions}} \in \mathbb{R}^{L \times (d+5)} \) 
encodes the sequence of event messages before \(t_i^{0}\). 
Each row corresponds to an action, described by its type (embedded in \(\mathbb{R}^d\)) and the five behavioral statistics 
of the submitting agent.  
This sequence is processed by the Action KAN-Transformer to produce action sequence embeddings:
\[
\mathbf{z}_{\text{a}} = \text{KAN-Transformer}(\mathbf{A}_{\text{actions}}).
\]

\subsubsection{Combining LOB, agent actions, and queue position}

The outputs of the two encoders are concatenated with the queue position \(q\) 
of the studied order, which reflects its execution priority under price–time matching:
\[
\mathbf{z}_{\text{c}} = \begin{bmatrix}
\mathbf{z}_{\text{a}} \\
\mathbf{z}_{\text{L}} \\
queue
\end{bmatrix}.
\]

\subsubsection{Predictor}

The {Predictor} maps the combined representation \( \mathbf{z}_{\text{c}} \) 
to compute the survival function \( S(t) \) and the density function \( f(t) \), 
which together determine the likelihood of execution at different horizons. Specifically, we assume that the time-to-fill \( T \) of a limit order follows a Weibull distribution\textsuperscript{3}. While KANFormer uses KAN-based encoders for representation learning, the predictor assumes a parametric Weibull distribution. {Monotonicity of the survival function is guaranteed, ensuring that \( S(t) \) is non-increasing, as required in survival analysis.
}
\footnotetext[3]{Empirically, we found Weibull fit stable and competitive; exploring richer parametric families or discrete-time hazards is left for future work.}

\[
T \mid \mathbf{z}_{\text{c}} \sim 
\text{Weibull}\!\left(\lambda(\mathbf{z}_{\text{c}}), \, 
k(\mathbf{z}_{\text{c}})\right),
\] 
where \( \lambda(\mathbf{z}_{\text{c}}) > 0 \) is the {scale parameter} 
and \( k(\mathbf{z}_{\text{c}}) > 0 \) is the {shape parameter}.  
Both parameters are predicted as functions of \( \mathbf{z}_{\text{c}} \). 
In practice, we compute \( \log \lambda \) and \( \log k \) for numerical stability \citep{Monod2024}.  

The corresponding survival function is:
\[
S(t \mid \mathbf{z}_{\text{c}}) 
= \exp\!\left[-\left(\frac{t}{\lambda(\mathbf{z}_{\text{c}})}\right)^{k(\mathbf{z}_{\text{c}})}\right],
\]
which is non-increasing in \( t \), consistent with the definition of survival probability 
\citep{kleinbaum1996survival}.  
The associated density function is:
\[
f(t \mid \mathbf{z}_{\text{c}}) 
= \frac{k(\mathbf{z}_{\text{c}})}{\lambda(\mathbf{z}_{\text{c}})} 
\left(\frac{t}{\lambda(\mathbf{z}_{\text{c}})}\right)^{k(\mathbf{z}_{\text{c}})-1} 
\exp\!\left[-\left(\frac{t}{\lambda(\mathbf{z}_{\text{c}})}\right)^{k(\mathbf{z}_{\text{c}})}\right].
\]

\subsection{Training details}\label{sec:training_details}

\paragraph{Data splits}  
We analyze the limit order book of front-month CAC 40 index futures contracts. 
The dataset was provided by Euronext, spanning 300 consecutive trading days 
from January 6th, 2016 to March 7th, 2017. We focus on orders submitted between 
9:05 am and 5:00 pm each day. 
The dataset is divided chronologically into three subsets. For each subset, the data is generated using the protocol presented in Section~\ref{sec:data_generation}. This split ensures that evaluation is performed strictly out-of-sample.
\begin{itemize}
    \item {Training set}: January 6th, 2016 to December 12th, 2016, {used to fit the model parameters.}
    \item {Validation set}: December 13th, 2016 to January 24th, 2017, {used for hyperparameter tuning and early stopping.}
    \item {Test set}: January 25th, 2017 to March 7th, 2017, {held out for final performance evaluation.}
\end{itemize}

\paragraph{Hyperparameters}  
Models are trained with the Adam optimizer \citep{kingma2014adam}, using an initial learning rate of $10^{-3}$ and apply an exponential learning rate decay schedule with factor $\gamma = 0.9$ at each epoch. Training is performed on an NVIDIA RTX A6000 with 48 GB of GPU memory. The lookback window is fixed at $L=50$. Hyperparameters are selected by minimizing validation RCLL at 200 epochs with early stopping (patience 10). {More details on the grid search are provided in Appendix~\ref{appendix:hyperparameters}.}

\paragraph{Loss function}  
We train KANFormer by minimizing the negative right-censored log-likelihood (RCLL).

\begin{equation}
    \mathcal{L} = -\sum_{i=1}^n \Big[ \delta_i \log f(T^C_i \mid \mathbf{z}_{\text{c}}) + (1 - \delta_i) \log S(T^C_i \mid \mathbf{z}_{\text{c}}) \Big],
    \label{eq:log_likelihood1}
\end{equation}

\paragraph{Train, test, validation protocol}  
{To account for variability, all experiments were repeated 30 times. 
We generated 30 distinct training, validation, and test sets 
$\{\mathcal{D}_{\text{train}}^k, \mathcal{D}_{\text{validation}}^k, \mathcal{D}_{\text{test}}^k\}_{k=1}^{30}$ 
according to the protocol in Section~\ref{sec:data_generation}. 
For each split $k$, we trained a model $\mathcal{M}_k$ on $\mathcal{D}_{\text{train}}^k$, 
validated it on $\mathcal{D}_{\text{validation}}^k$, and evaluated it on $\mathcal{D}_{\text{test}}^k$ to obtain the four metrics. To ensure stable convergence and reduce computational cost, models $\mathcal{M}_2$ to $\mathcal{M}_{30}$ were \emph{warm-initialized} with the parameters of $\mathcal{M}_1$ rather than with random initialization. Warm initialization, which reuses weights trained on a related dataset, has been shown to accelerate optimization and stabilize training in deep learning \citep{ash2020warm, sambharya2024learning, wang2024flexnet}. This strategy ensures that variability across runs reflects only dataset composition, not random initialization effects. We therefore report all results as averages (with standard deviations) over the 30 experiments, 
providing a controlled quantification of variability in predictive performance.}

\subsection{Summary}

In summary, KANFormer integrates agent actions, LOB snapshots at the moments of those actions, 
and the queue position of the studied order into a unified Transformer-based framework 
enhanced with KANs. This design enables the model to capture market-level dynamics, the drivers of these dynamics and order-level priority, thereby ensuring a well-calibrated 
and monotonic survival function. In the next section, we evaluate KANFormer on CAC 40 futures data and benchmark it against existing approaches for fill probability estimation.

\section{Experiments and Results}\label{sec:experiments}

In this section, we empirically evaluate KANFormer on the CAC 40 futures dataset introduced in Section~\ref{sec:data_generation}. The objective is to assess the model’s ability to predict the survival function of limit orders and compare its performance against existing baselines. We begin by introducing the evaluation metrics and the benchmarks. We then present results from three perspectives: (i) Evaluation metrics and benchmarks, (ii) ablation studies that analyze the impact of input features (Section~\ref{sec:problem_formulation}) and architectural choices (Section~\ref{sec:model}), and (iii) feature attribution analysis, which provides a dynamic view of how feature importance evolves over time. 

\subsection{Evaluation metrics and baselines}\label{sec:metrics}

As introduced in Section~\ref{sec:survival_analysis}, we evaluate models using both 
\textit{calibration} and \textit{discrimination} metrics. 
Calibration assesses whether predicted survival probabilities align with observed event frequencies, 
while discrimination measures how well the model ranks orders by their probabilities of execution. Together, these complementary perspectives provide a balanced assessment of survival prediction quality. We adopt four widely used metrics, {here we only present the intuition of theses metrics, formal definitions are provided in Appendix~\ref{appendix:metrics}.}

\paragraph{Negative Right-Censored Log-Likelihood (RCLL)}  
RCLL \citep{balakrishnan2012left, rindt2022survival} evaluates the fit of the predicted survival and density functions under censoring.  
Importantly, RCLL is a \textit{proper scoring rule} (see Appendix~\ref{appendix:proper_scoring} for a formal definition), 
which means that in expectation it is uniquely minimized by the true survival distribution. \textit{Intuition:} A lower RCLL indicates better calibration, i.e., the predicted distributions match the observed outcomes more closely.  

\paragraph{Concordance Index (C-index)}  
We use the \emph{classical}, time-independent C-index \citep{harrell1996multivariable, uno2011c}.  
It provides a global measure of ranking quality by comparing all pairs of orders: 
a pair is concordant if the order with a higher predicted risk executes earlier. 
Pairs are comparable only if the earlier event is observed (not censored). In our experiments, we compute a fixed risk score per order as $r_i = 1 - \hat{S}(t^\ast \mid \mathbf{x}_i)$, where $t^\ast$ is the largest evaluation horizon so that $r_i$ aggregates risk over the evaluation window. \textit{Intuition:} The C-index summarizes overall discrimination across the entire time axis, complementing the horizon-specific information of the AUC. 

\paragraph{Time-dependent Brier Score}  
The Brier score measures the squared error between predicted survival probabilities and observed outcomes at time \(t\) \citep{lee2019dynamic, graf1999assessment}.  
\textit{Intuition:} It reflects both calibration of predictions; lower values indicate more accurate survival curves.  Also, since BS is time-dependent, we report the Integrated Brier Score (IBS) across 20 evaluation horizons chosen between the 10th and 50th percentiles (upper bound is 0.627 seconds) of the event time distribution: {$IBS = \frac{1}{\tau_2 - \tau_1} \int_{\tau_1}^{\tau_2} BS(t)\,dt$}, where \(\tau_1\) and \(\tau_2\) denote the lower and upper bounds of the evaluation window.

\paragraph{Time-dependent AUC}  
The AUC \citep{lambert2016summary} quantifies the model’s ability to discriminate between 
orders executed by time \(t\) (cases) and those still active (controls). Like IBS, we report the average across 20 evaluation horizons chosen between the 10th and 50th percentiles (upper bound is 0.627 seconds) of the event time distribution: {$IAUC = \frac{1}{\tau_2 - \tau_1} \int_{\tau_1}^{\tau_2} AUC(t)\,dt$}, where \(\tau_1\) and \(\tau_2\) denote the lower and upper bounds of the evaluation window.
\textit{Intuition:} A higher AUC at horizon \(t\) means executed orders are consistently assigned higher risk than unexecuted ones.

\paragraph{Baselines}  
Following \citet{arroyo2024deep}, we benchmark our model against MLP, LSTM \citep{hochreiter1997long}, DeepHit \citep{lee2018deephit}, and LSTM\_Hazard \citep{maglaras2022deep}.  
Note that DeepHit treats survival time as discrete; we linearly interpolate the discrete CDF between bin midpoints to obtain continuous \(S(t)\) and \(f(t)\). {We denote the convolutional-transformer model of \citet{arroyo2024deep} as ConvTrans, which uses only $\mathbf{X}_{\text{LOB}}$; ConvTrans$^{++}$ extends this baseline with action-type embeddings, agent features, and queue position.}  
We also include two classical survival models: Random Forests (RF) \citep{ishwaran2008random} and the Cox model \citep{simon2011regularization}, both implemented with the \texttt{scikit-survival} package \citep{polsterl2020scikit}.

\subsection{Results of experiments}

\begin{table}[t]
    \centering
    \begin{tabular}{cccccc}
        \hline
       \multicolumn{2}{c}{Model} & RCLL ($\downarrow$) & IBS ($\downarrow$) & IAUC ($\uparrow$) & C-index ($\uparrow$) \\
        \hline
        \multirow{2}{*}{Non-deep} 
            & RF  & 1.64 $\pm$ 0.06  & \textbf{0.027 $\pm$ 0.002} & 0.64 $\pm$ 0.02 & 0.63 $\pm$ 0.04 \\
            & Cox & 1.74 $\pm$ 0.05  & \textbf{0.027 $\pm$ 0.002} & 0.68 $\pm$ 0.02 & 0.67 $\pm$ 0.05 \\
        \hline
        \multicolumn{2}{c}{MLP}          & 1.89 $\pm$ 0.20 & 0.029 $\pm$ 0.003 & 0.53 $\pm$ 0.04 & 0.51 $\pm$ 0.04 \\
        \hline
        \multicolumn{2}{c}{LSTM}         & 1.33 $\pm$ 0.20 & 0.029 $\pm$ 0.003 & 0.58 $\pm$ 0.03 & 0.52 $\pm$ 0.03 \\
        \multicolumn{2}{c}{LSTM\_Hazard} & 1.64 $\pm$ 0.06 & 0.028 $\pm$ 0.003 & 0.54 $\pm$ 0.02 & 0.52 $\pm$ 0.03 \\
        \hline
        \multicolumn{2}{c}{ConvTrans}     & 1.18 $\pm$ 0.06 & 0.028 $\pm$ 0.002 & 0.47 $\pm$ 0.04 & 0.44 $\pm$ 0.06 \\
        \multicolumn{2}{c}{ConvTrans$^{++}$} & 0.93 $\pm$ 0.05 & 0.028 $\pm$ 0.003 & 0.54 $\pm$ 0.04 & 0.49 $\pm$ 0.05 \\
        \hline
        \multicolumn{2}{c}{DeepHit}      & 0.56 $\pm$ 0.03 & 0.028 $\pm$ 0.002 & 0.56 $\pm$ 0.04 & 0.58 $\pm$ 0.04 \\
        \hline
        \multicolumn{2}{c}{KANFormer}    & \textbf{0.53 $\pm$ 0.03} & \textbf{0.027 $\pm$ 0.001} & \textbf{0.76 $\pm$ 0.02} & \textbf{0.72 $\pm$ 0.05} \\
        \hline
    \end{tabular}
    \caption{Comparison of model performance on the test set. Apart from ConvTrans, all other models are trained on the enriched input. RCLL and IBS are reported as calibration metrics, while AUC and C-index serve as discrimination metrics. 
For AUC and BS, results are averaged over 20 evaluation horizons between the 10th and 50th percentiles (upper bound is 0.627 seconds) of the event-time distribution. 
The C-index is the classical, time-independent version, assessing global ranking quality across all pairs of orders. {Reported errors correspond to standard deviations across 30 experiments.}}
    \label{table:impacts_of_models}
\end{table}

\begin{figure}[t]
    \centering
    \includegraphics[width=1.0\textwidth]{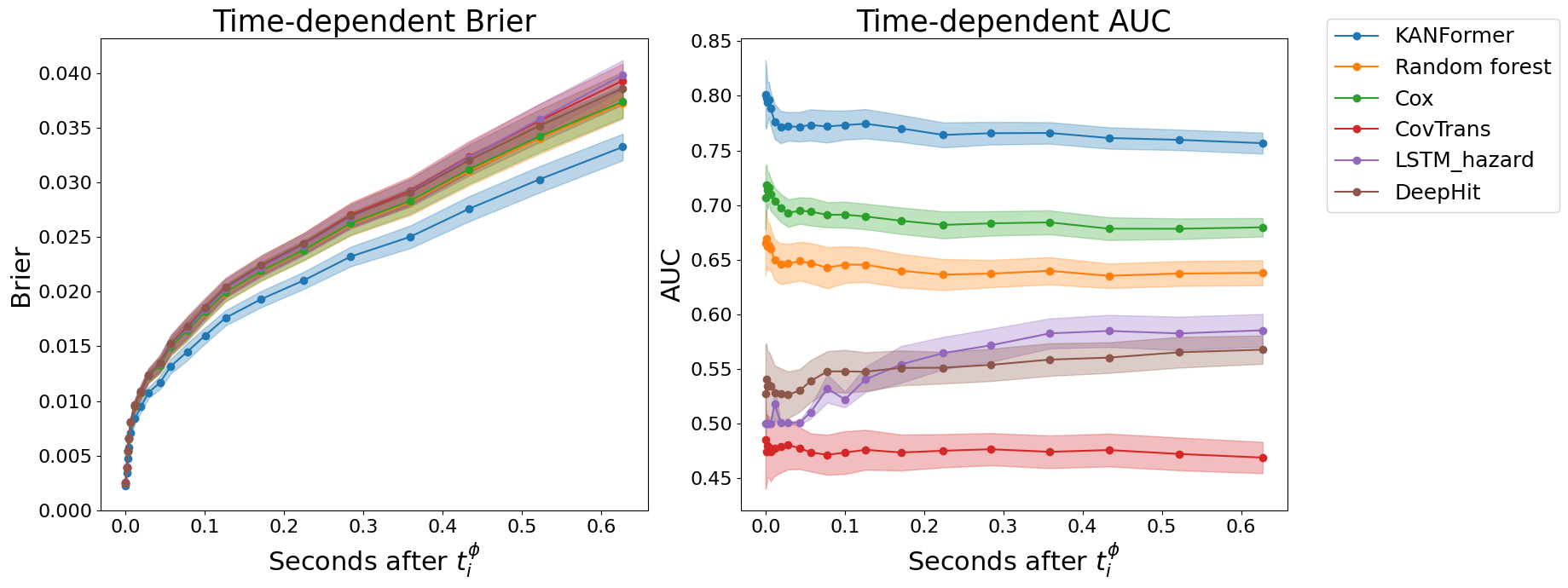}
    \caption{The evolution of Brier score and AUC over a set of 20 different prediction time points.}
    \label{Fig.Evolution_metrics}
\end{figure}

Table~\ref{table:impacts_of_models} summarizes the performance of various models in the limit order book setting. For ConvTrans, only $\mathbf{X}_{\text{LOB}}$ is used. All other models, including ConvTrans$^{++}$, incorporate action type embeddings, agent features, and queue position. Figure~\ref{Fig.Evolution_metrics} presents the evolution of evaluation metrics over time. Our model outperforms all baselines across all metrics. In terms of calibration, it achieves the lowest RCLL (0.53) and IBS (0.027), indicating highly accurate estimation of the survival distribution. For discrimination, it obtains the highest IAUC (0.76) and C-index (0.72), reflecting superior ability to rank execution risks.

Compared with non-deep models such as Cox and Random Forests, our approach yields substantial improvements in IAUC and C-index, along with better calibration. Among deep learning baselines, ConvTrans, using only $\mathbf{X}_{\text{LOB}}$, performs poorly, with near-random discrimination (IAUC and C-index around 0.47). However, ConvTrans$^{++}$ shows marked improvements, confirming the importance of including agent-level information and queue position. DeepHit attains relatively strong calibration (RCLL = 0.56, IBS = 0.028), though not as good as KANFormer, but its discrimination is limited (IAUC = 0.56, C-index = 0.58), likely due to its discretized output. In contrast, our model achieves both precise distribution estimation and robust ranking. This demonstrates that accurately modeling fine-grained order dynamics, enriched with behavioral features and queue context, and preserving monotonicity in the survival function, is key to achieving state-of-the-art fill probability estimation.

 \subsection{Ablation study}
  
  We now conduct an ablation study to show the impact of model structure and the input features. 
 
 \subsubsection{Impact of model structure}

Table \ref{tab:impact_of_structure} presents an ablation study evaluating the contributions of Kolmogorov–Arnold Networks (KAN) and dilated causal convolution (DCC) to model performance. Across all metrics, the results consistently highlight the benefits of both components.

First, comparing KANFormer to its Transformer counterpart, we observe that replacing standard feedforward layers with KAN layers improves both calibration and discrimination. Without convolution, KANFormer improves IAUC from 0.61 to 0.71 and C-index from 0.54 to 0.69. The performance improves further when DCC is added: IAUC increases from 0.63 (Transformer + DCC) to 0.76 (KANFormer + DCC), and C-index from 0.56 to 0.72. These findings support the hypothesis that KAN's spline-based functional decomposition enhances representation capacity, especially in modeling complex nonlinear dynamics in high-frequency financial environments.

Second, the inclusion of DCC-based convolution improves performance in all cases, demonstrating the importance of capturing local temporal patterns in the limit order book. For KANFormer, adding DCC improves both calibration (RCLL 0.53 to 0.52) and discrimination (IAUC 0.71 to 0.76; C-index 0.69 to 0.72). For the Transformer backbone, DCC primarily improves discrimination (IAUC 0.61 to 0.63; C-index 0.54 to 0.56), while calibration changes little.

\begin{figure}[t]
 \centering
    \includegraphics[width=1\textwidth]{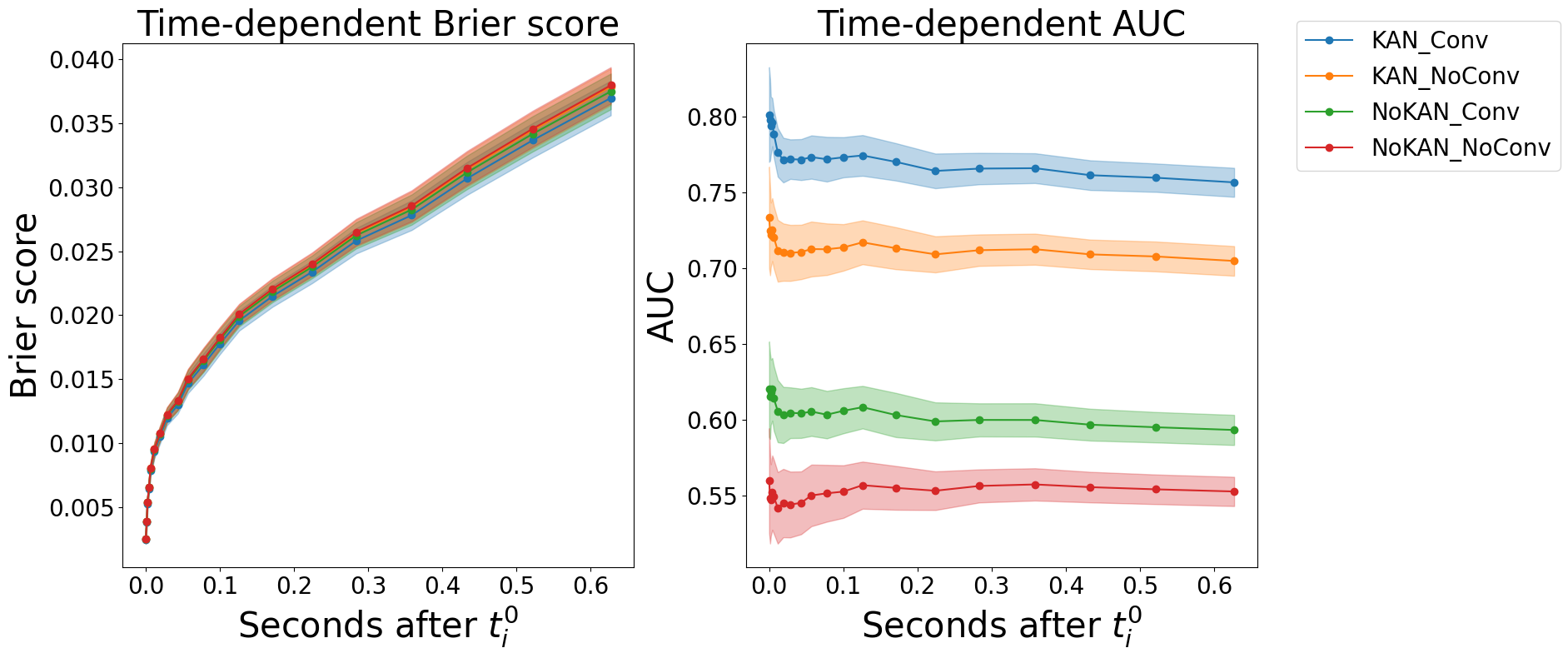}
    \caption{The evolution of metrics over a set of 20 prediction time points.}
\label{Fig.Evolution_metrics1}
\end{figure}

Overall, this ablation confirms that the combination of KAN and DCC leads to optimal performance. DCC contributes effective local temporal context, while KAN enhances the functional approximation capacity of the feedforward blocks within the encoders. Together, they result in a highly expressive and well-calibrated survival model tailored for high-frequency trading applications. As in the previous subsection, we also report the evolution of the Brier Score and AUC in Figure~\ref{Fig.Evolution_metrics1}.

 \begin{table}[t]
\centering

\begin{tabular}{cccccc}
\hline
 \multicolumn{2}{c}{Model} & RCLL ($\downarrow$) & IBS ($\downarrow$) & IAUC ($\uparrow$)  & C-index ($\uparrow$) \\
\hline
\multirow{2}{*}{KANFormer} 
    & DCC     & \textbf{0.53 $\pm$ 0.03} & \textbf{0.027 $\pm$ 0.001} & \textbf{0.76 $\pm$ 0.02} & \textbf{0.72 $\pm$ 0.05} \\
    & No DCC  & \textbf{0.53 $\pm$ 0.03}          & \textbf{0.027 $\pm$ 0.002}          & 0.71 $\pm$ 0.02          & 0.69 $\pm$ 0.04 \\
\hline
\multirow{2}{*}{Transformer} 
    & DCC     & 0.56 $\pm$ 0.03          & \textbf{0.027 $\pm$ 0.002}          & 0.63 $\pm$ 0.04          & 0.56 $\pm$ 0.04 \\
    & No DCC  & 0.57 $\pm$ 0.02          & \textbf{0.027 $\pm$ 0.002}          & 0.61 $\pm$ 0.02          & 0.54 $\pm$ 0.05 \\
\hline
\end{tabular}
\caption{Ablation study: impact of KAN and DCC components on model performance. 
All configurations use the enriched input consisting of action-type embeddings, agent features, and queue position. 
RCLL and IBS are reported as calibration metrics, while AUC and C-index serve as discrimination metrics. 
For AUC and BS, results are averaged over 20 evaluation horizons between the 10th and 50th percentiles (upper bound is 0.627 seconds) of the event-time distribution. 
The C-index is the classical, time-independent version. {Reported errors correspond to standard deviations across 30 experiments.}}
\label{tab:impact_of_structure}
\end{table}

\subsubsection{Impact of input features}\label{sec:Impact_features}

As introduced in Section~\ref{sec:data_generation}, compared to the work of \citet{arroyo2024deep, maglaras2022deep}, we incorporate additional features into our model, including agent actions (represented by action type embeddings and five representative statistics per agent) and the queue position of the target order. To evaluate the contribution of each feature, we conduct an ablation study using the KANFormer + DCC configuration.

Table~\ref{tab:results_summary_yes_no} summarizes the results. In the table, ‘yes’ indicates that a feature is included in the model, while ‘no’ indicates its exclusion. The findings clearly show that each feature contributes meaningfully to predictive performance, with the best results achieved when ActionType, Agent Features, and Queue are included.

Excluding the queue information leads to the most substantial decline in performance: IAUC drops from 0.76 to 0.57 and C-index from 0.72 to 0.55. This underscores the critical role of queue position in predicting time-to-fill of limit orders, as it reflects both priority in the order book and local liquidity pressure, two factors essential to accurate fill probability modeling.

Removing agent features also results in notable degradation, albeit less pronounced. IAUC falls to 0.68 and C-index to 0.65, suggesting that agent-level statistics (e.g., Market Ratio, Cancel Ratio) provide valuable insights into the behavioral patterns of market participants. {It is important to note that the behavioral statistics of agents, as used in this work, are not observable to a single market participant. Therefore, an individual agent cannot directly implement the full model described in this paper.}

Finally, omitting ActionType (while retaining Agent Features and Queue) yields lower performance (IAUC = 0.66, C-index = 0.62). This result indicates that the nature of actions such as insertions, cancellations, or modifications contributes to modeling the evolving dynamics and intentions within the LOB.

\begin{figure}[t]
 \centering
    \includegraphics[width=1\textwidth]{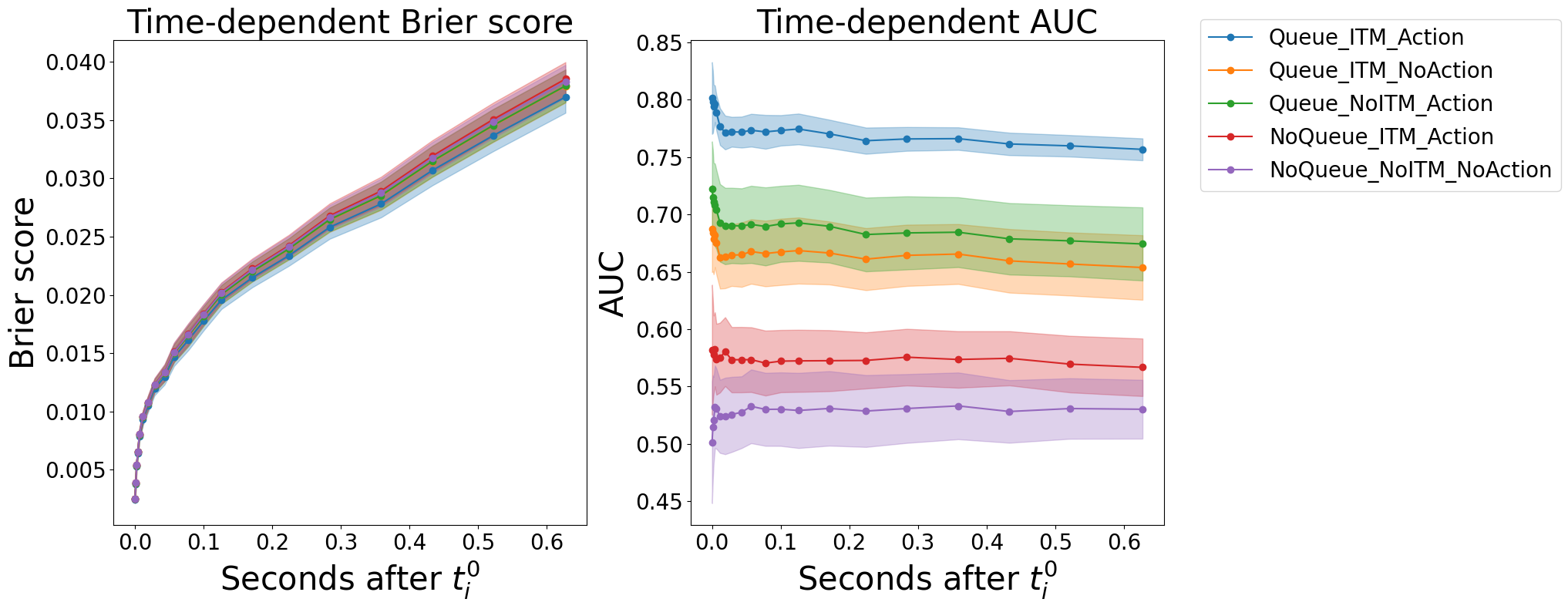}
    \caption{The evolution of metrics over a set of 20 prediction time points.}
\label{Fig.Evolution_metrics2}
\end{figure}

Overall, the ablation study confirms that ActionType, Agent Features, and Queue provide complementary and essential information for effective survival analysis of limit orders. This conclusion is also supported by Table~\ref{table:impacts_of_models}, {it can be seen that} the extended model $ConvTrans^{++}$ outperforms the baseline ConvTrans across all metrics. As in the previous subsection, we also report the evolution of the Brier Score and AUC in Figure~\ref{Fig.Evolution_metrics2}.

\begin{table}[t]
\centering
\begin{tabular}{ccccccc}
\hline
ActionType & AgentFeatures & Queue & RCLL ($\downarrow$) & IBS ($\downarrow$) & IAUC ($\uparrow$) & C-index ($\uparrow$) \\
\hline
yes & yes & yes & \textbf{0.53 $\pm$ 0.03} & \textbf{0.027 $\pm$ 0.001} & \textbf{0.76 $\pm$ 0.02} & \textbf{0.72 $\pm$ 0.05} \\
yes & yes & no  & 0.54 $\pm$ 0.03 & 0.028 $\pm$ 0.002 & 0.57 $\pm$ 0.07 & 0.55 $\pm$ 0.05 \\
yes & no  & yes & 0.53 $\pm$ 0.02 & \textbf{0.027 $\pm$ 0.002} & 0.68 $\pm$ 0.06 & 0.65 $\pm$ 0.06 \\
no  & yes & yes & 0.53 $\pm$ 0.03 & \textbf{0.027 $\pm$ 0.001} & 0.66 $\pm$ 0.06 & 0.62 $\pm$ 0.05 \\
no  & no  & no  & 0.54 $\pm$ 0.02 & 0.028 $\pm$ 0.003 & 0.53 $\pm$ 0.07 & 0.52 $\pm$ 0.06 \\
\hline
\end{tabular}
\caption{Ablation study on input features using the KANFormer+DCC configuration. 
The table reports performance when including or excluding ActionType, Agent Features, and Queue position as inputs. 
A value of ‘yes’ indicates inclusion of the feature, while ‘no’ indicates exclusion. 
RCLL and IBS are reported as calibration metrics, while AUC and C-index serve as discrimination metrics. 
For AUC and BS, results are averaged over 20 evaluation horizons between the 10th and 50th percentiles (upper bound is 0.627 seconds) of the event-time distribution. 
The C-index is the classical, time-independent version. {Reported errors correspond to standard deviations across 30 experiments.}}
\label{tab:results_summary_yes_no}
\end{table}

\subsection{Feature contribution}

\begin{figure}[t]
\centerline{\includegraphics[scale=0.60]{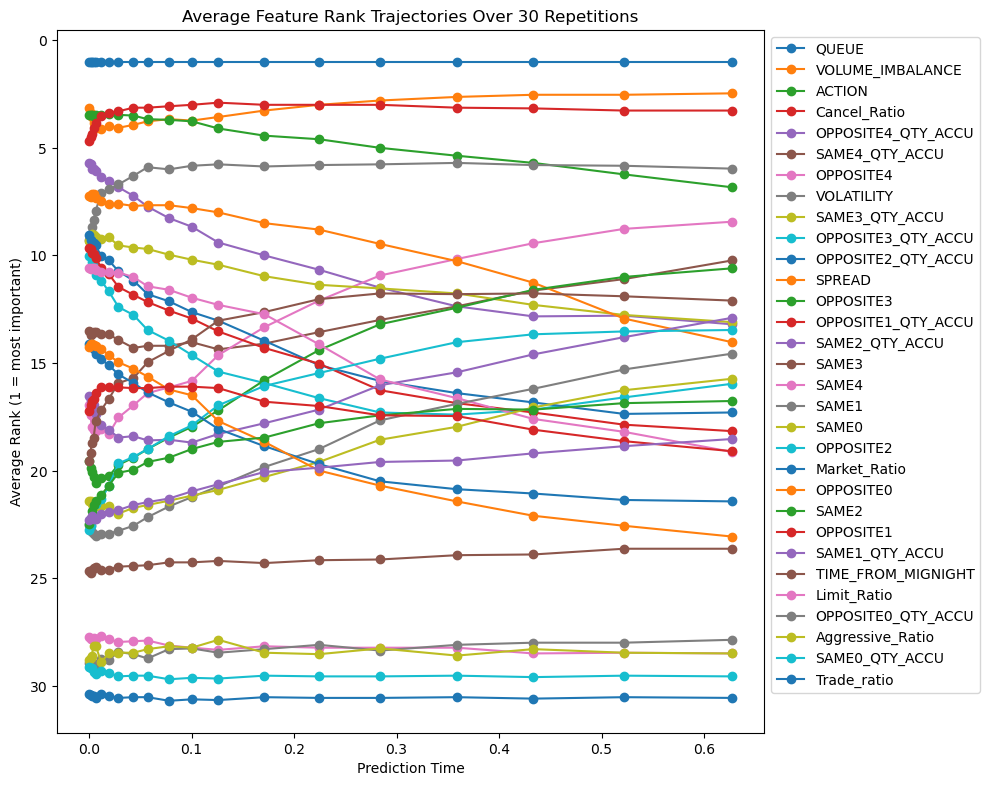}}
\caption{Evolution of feature importance of different prediction horizons using our model. The SHAP values are averaged over all orders in the test set.}
\label{fig:evolution_metrics_all}
\end{figure}

In Section~\ref{sec:Impact_features}, {it has been} demonstrated that incorporating ActionType, Agent Features, and Queue information enhances survival analysis for limit orders. While the previous analysis adopted a modular approach by training separate models with different feature combinations, we now focus on a fixed configuration: KANFormer combined with DCC, using all three feature types. Our goal is to illustrate how these features contribute to the model’s predictions over varying time horizons.

To this end, we employ SHAP (SHapley Additive exPlanations) for feature importance visualization \citep{lundberg2017unified}. Specifically, SHAP unifies several existing methods under a consistent framework that satisfies desirable properties such as local accuracy, missingness, and consistency \citep{lundberg2017unified}. By assigning each feature an importance value representing its marginal contribution to the model prediction, SHAP enables a comprehensive and theoretically grounded understanding of complex model behavior. While KernelSHAP, a widely used variant, provides model-agnostic explanations by estimating Shapley values through sampling, it is computationally intensive, especially for deep learning models with large input dimensions. To address this limitation, we adopt the GradientExplainer provided by the SHAP library, which leverages model gradients to efficiently approximate Shapley values. GradientExplainer is specifically designed for interpreting deep learning models, offering a more scalable and practical solution. We leverage GradientExplainer to quantify and visualize the relative importance of features within our KANFormer+DCC model.

Figure~\ref{fig:evolution_metrics_all} illustrates the evolution of feature importance across various prediction horizons. Notably, the queue position of the studied order emerges as the most influential feature in the short-term horizon, reaffirming its central role in execution modeling, as previously emphasized in the queue-reactive model of \citet{huang2015simulating} and the microstructural analysis by \citet{cartea2014buy}. This is intuitive, as an order’s position in the queue directly determines its priority and likelihood of being matched, especially within short intervals where market conditions are relatively stable. 

Volume imbalance also shows consistently high importance across different horizons. As a fast-reacting variable, it encapsulates real-time shifts in demand and supply on both sides of the book. Its predictive power lies in its capacity to capture transient liquidity imbalances, which often precede aggressive trades or cancellations. This aligns with the findings of \citet{arroyo2024deep}, where volume imbalance was also identified as a top-performing feature in dynamic execution modeling. 

\begin{figure}[t]
\centering

\begin{minipage}{0.8\textwidth}
    \centering
    \includegraphics[width=\linewidth]{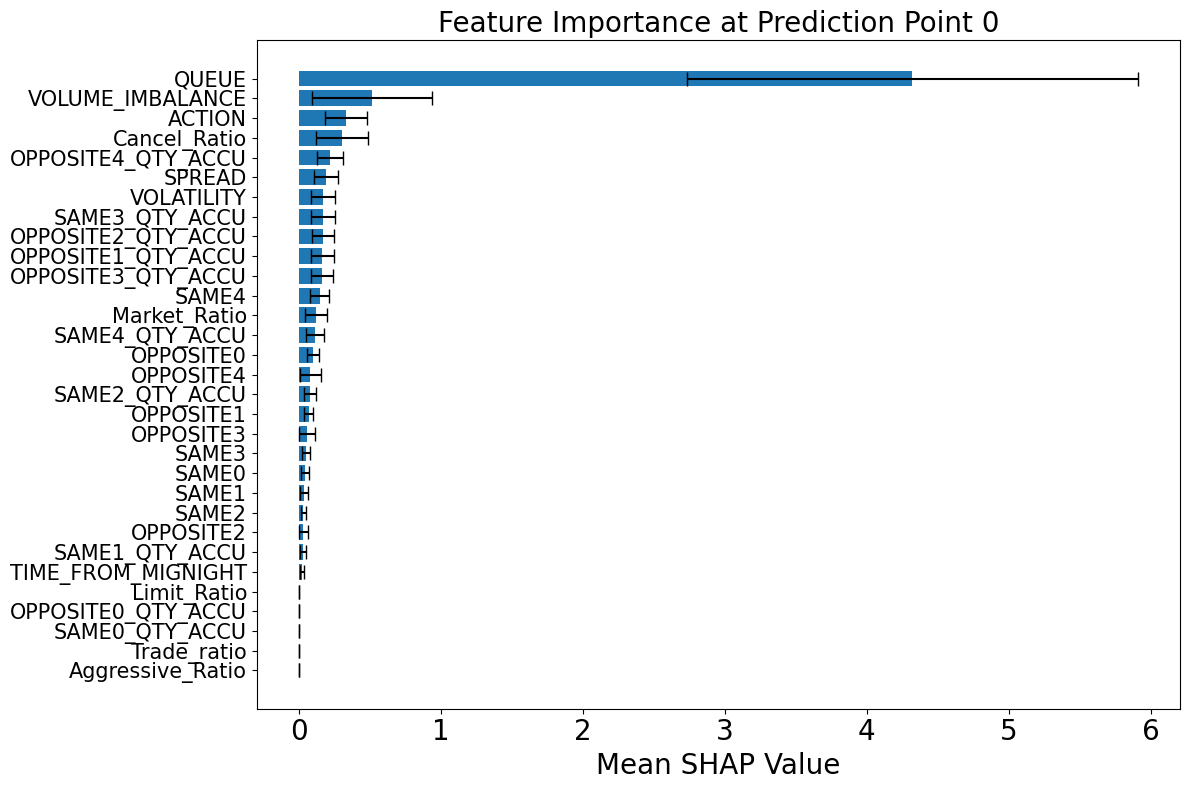}
    \vspace{3pt}
    {\small (a) Feature importance at 1.3 ms}
\end{minipage}

\begin{minipage}{0.8\textwidth}
    \centering
    \includegraphics[width=\linewidth]{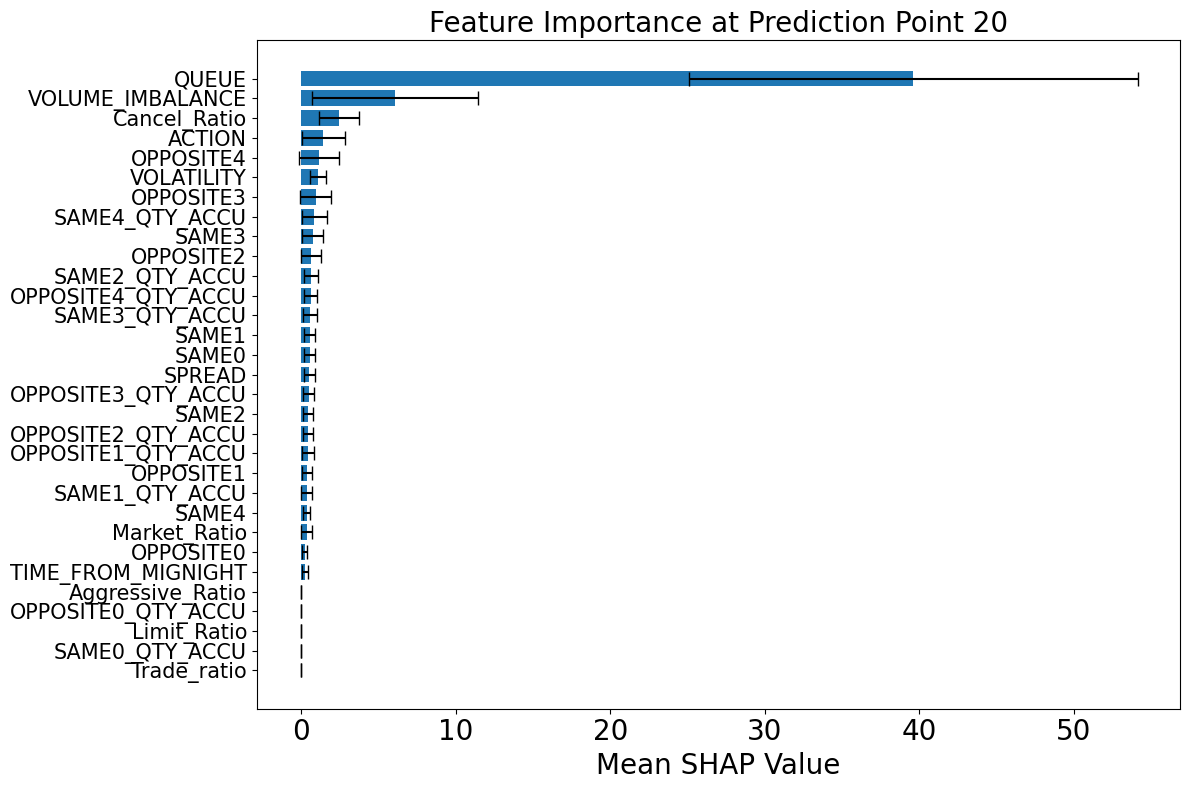}
    {\small (b) Feature importance at 0.627 s}
\end{minipage}

\caption{Feature importance at two representative time points: 
1.3 ms (upper) and 0.627 seconds (lower).}
\label{fig:lob1}
\end{figure}

The type of action initiated by agents, whether it be an insertion, cancellation, or modification, gains relevance as the prediction horizon lengthens. This trend reflects the growing impact of behavioral signals on execution risk over time. For example, a surge in cancellation actions may indicate a shift in market sentiment or deteriorating order book depth, both of which can adversely affect the fill probability of a pending order. As discussed in Section~\ref{sec:LOBRE}, these action types serve as early indicators of directional changes and liquidity shifts.

Among the agent-specific features, the cancel ratio consistently ranks as one of the most informative variables. This metric encapsulates the behavioral tendencies of market participants and helps distinguish between liquidity providers and fleeting liquidity. A high cancel ratio typically signals less commitment to maintaining liquidity, thereby increasing the uncertainty around execution outcomes. Its importance in predicting fill probabilities underlines the value of integrating long-term agent behavior into execution models.

Figure~\ref{fig:lob1} provides a more granular view of SHAP values at two representative prediction times. At earlier time points (e.g., 1.3 ms), the model relies heavily on the queue position and volume imbalance, reflecting a focus on immediate liquidity and priority dynamics. At later prediction points (e.g., 0.627 seconds), the importance of action type and cancel ratio increases, suggesting that the model shifts attention from microstructural immediacy to behavioral messages that reflect evolving market intentions. Collectively, these results demonstrate that the relative importance of predictive features is not static but evolves with the prediction horizon, which is not studied by previous works.

\begin{figure}[t]
\centerline{\includegraphics[scale=0.60]{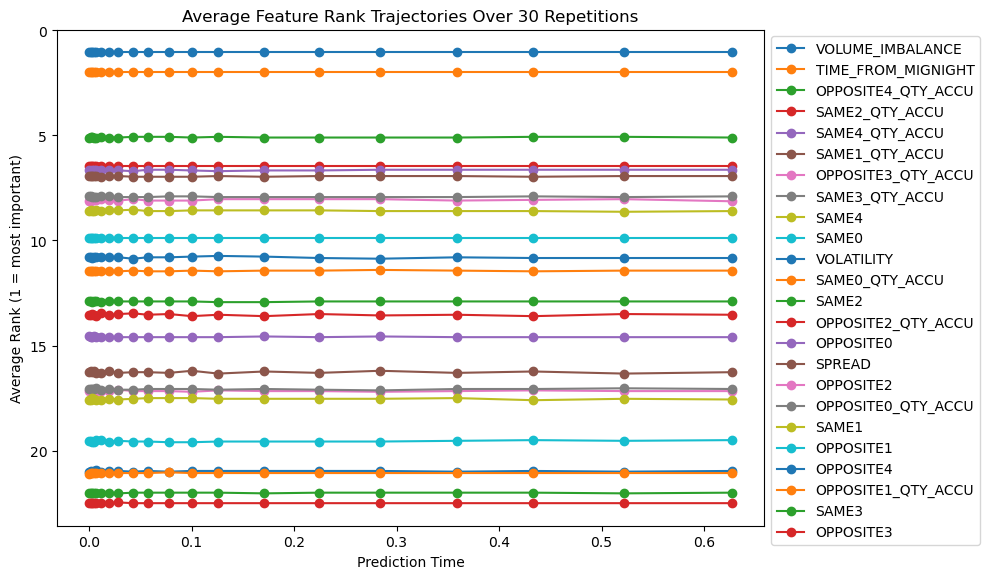}}
\caption{Evolution of feature importance of different prediction horizons using the ConvTrans model \citep{arroyo2024deep}. The SHAP values are averaged over all orders in the test set.}
\label{fig:evolution_metrics_all_1}
\end{figure}

For comparison, Figure~\ref{fig:evolution_metrics_all_1} shows the evolution of feature importance according to the model proposed by \citet{arroyo2024deep}, where fast-moving features, such as volume imbalance, are consistently assigned the highest importance.

\section{Conclusions and perspectives}\label{sec:conclusions_perspectives}

In this paper, we introduced a new deep survival model specifically designed for predicting the execution risk of limit orders in electronic financial markets. Building on a rigorous survival analysis framework, our model integrates not only the traditional features of the evolving limit order book but also agent-level behavioral features, which capture the dynamics of order flows more comprehensively. Furthermore, by incorporating Kolmogorov–Arnold Networks (KANs) into the Transformer architecture, we enhanced the expressiveness of the model, resulting in improved predictive performance. Our evaluation relied on both discrimination and calibration metrics, providing a holistic assessment of the model’s reliability. Finally, we utilized SHAP to analyze feature importance over time.

Several directions for future research emerge from this work. First, while we have focused on data from the Euronext front-month CAC 40 index futures, it would be valuable to apply our framework to individual equities across different sectors to investigate how execution likelihood patterns vary between asset classes. Such an extension could offer insights into sector-specific market microstructures. Second, while our model directly parameterizes a monotonically decreasing survival function, alternative approaches exist, such as modeling the probability mass function (PMF) or the discrete-time hazard rate \citep{kvamme2021continuous}. Exploring these discrete-time modeling strategies, along with their discretization schemes and associated trade-offs, represents a promising avenue for further development.

\section*{Acknowledgements}
This work has received support from the French government, managed by the National Research Agency (ANR), 
under the ``France 2030'' program with reference ANR-23-IACL-0008.

\bibliographystyle{abbrvnat}
\bibliography{references}

@article{cho2000probability,
  title={The probability of limit-order execution},
  author={Cho, Jin-Wan and Nelling, Edward},
  journal={Financial Analysts Journal},
  volume={56},
  number={5},
  pages={28--33},
  year={2000},
  publisher={Taylor \& Francis}
}

@article{huang2015simulating,
  title={Simulating and analyzing order book data: The queue-reactive model},
  author={Huang, Weibing and Lehalle, Charles-Albert and Rosenbaum, Mathieu},
  journal={Journal of the American Statistical Association},
  volume={110},
  number={509},
  pages={107--122},
  year={2015},
  publisher={Taylor \& Francis}
}

@article{cartea2014buy,
  title={Buy low, sell high: A high frequency trading perspective},
  author={Cartea, {\'A}lvaro and Jaimungal, Sebastian and Ricci, Jason},
  journal={SIAM Journal on Financial Mathematics},
  volume={5},
  number={1},
  pages={415--444},
  year={2014},
  publisher={SIAM}
}

@article{Monod2024,
    doi = {10.21105/joss.07341},
    url = {https://doi.org/10.21105/joss.07341},
    year = {2024},
    publisher = {The Open Journal},
    volume = {9},
    number = {104},
    pages = {7341},
    author = {Mélodie Monod and Peter Krusche and Qian Cao and Berkman Sahiner and Nicholas Petrick and David Ohlssen and Thibaud Coroller},
    title = {TorchSurv: A Lightweight Package for Deep Survival Analysis}, journal = {Journal of Open Source Software}
}

@article{kingma2014adam,
  title={Adam: A method for stochastic optimization},
  author={Kingma, Diederik P and Ba, Jimmy},
  journal={arXiv preprint arXiv:1412.6980},
  year={2014}
}

@article{wu2019queue,
  title={Queue-reactive Hawkes models for the order flow},
  author={Wu, Peng and Rambaldi, Marcello and Muzy, Jean-Fran{\c{c}}ois and Bacry, Emmanuel},
  journal={arXiv preprint arXiv:1901.08938},
  year={2019}
}

@article{uno2011c,
  title={On the C-statistics for evaluating overall adequacy of risk prediction procedures with censored survival data},
  author={Uno, Hajime and Cai, Tianxi and Pencina, Michael J and D'Agostino, Ralph B and Wei, Lee-Jen},
  journal={Statistics in medicine},
  volume={30},
  number={10},
  pages={1105--1117},
  year={2011},
  publisher={Wiley Online Library}
}

@article{lee2019dynamic,
  title={Dynamic-deephit: A deep learning approach for dynamic survival analysis with competing risks based on longitudinal data},
  author={Lee, Changhee and Yoon, Jinsung and Van Der Schaar, Mihaela},
  journal={IEEE Transactions on Biomedical Engineering},
  volume={67},
  number={1},
  pages={122--133},
  year={2019},
  publisher={IEEE}
}

@article{zhang2019deeplob,
  title={Deeplob: Deep convolutional neural networks for limit order books},
  author={Zhang, Zihao and Zohren, Stefan and Roberts, Stephen},
  journal={IEEE Transactions on Signal Processing},
  volume={67},
  number={11},
  pages={3001--3012},
  year={2019},
  publisher={IEEE}
}

@article{LO200231,
title = {Econometric models of limit-order executions},
journal = {Journal of Financial Economics},
volume = {65},
number = {1},
pages = {31-71},
year = {2002},
issn = {0304-405X},
doi = {https://doi.org/10.1016/S0304-405X(02)00134-4},
url = {https://www.sciencedirect.com/science/article/pii/S0304405X02001344},
author = {Andrew W. Lo and A.Craig MacKinlay and June Zhang}
}

@article{kamarudin2017time,
  title={Time-dependent ROC curve analysis in medical research: current methods and applications},
  author={Kamarudin, Adina Najwa and Cox, Trevor and Kolamunnage-Dona, Ruwanthi},
  journal={BMC medical research methodology},
  volume={17},
  number={1},
  pages={53},
  year={2017},
  publisher={Springer}
}

@article{bleistein2024dynamical,
  title={Dynamical survival analysis with controlled latent states},
  author={Bleistein, Linus and Nguyen, Van-Tuan and Fermanian, Adeline and Guilloux, Agathe},
  journal={arXiv preprint arXiv:2401.17077},
  year={2024}
}

@article{gould2013limit,
  title={Limit order books},
  author={Gould, Martin D and Porter, Mason A and Williams, Stacy and McDonald, Mark and Fenn, Daniel J and Howison, Sam D},
  journal={Quantitative Finance},
  volume={13},
  number={11},
  pages={1709--1742},
  year={2013},
  publisher={Taylor \& Francis}
}

@article{arroyo2024deep,
  title={Deep attentive survival analysis in limit order books: Estimating fill probabilities with convolutional-transformers},
  author={Arroyo, Alvaro and Cartea, Alvaro and Moreno-Pino, Fernando and Zohren, Stefan},
  journal={Quantitative Finance},
  volume={24},
  number={1},
  pages={35--57},
  year={2024},
  publisher={Taylor \& Francis}
}

@article{maglaras2022deep,
  title={A deep learning approach to estimating fill probabilities in a limit order book},
  author={Maglaras, Costis and Moallemi, Ciamac C and Wang, Muye},
  journal={Quantitative Finance},
  volume={22},
  number={11},
  pages={1989--2003},
  year={2022},
  publisher={Taylor \& Francis}
}

@article{hung2010estimation,
  title={Estimation methods for time-dependent AUC models with survival data},
  author={Hung, Hung and Chiang, Chin-Tsang},
  journal={Canadian Journal of Statistics},
  volume={38},
  number={1},
  pages={8--26},
  year={2010},
  publisher={Wiley Online Library}
}

@book{kleinbaum1996survival,
  title={Survival analysis a self-learning text},
  author={Kleinbaum, David G and Klein, Mitchel},
  year={1996},
  publisher={Springer}
}

@article{leung1997censoring,
  title={Censoring issues in survival analysis},
  author={Leung, Kwan-Moon and Elashoff, Robert M and Afifi, Abdelmonem A},
  journal={Annual review of public health},
  volume={18},
  number={1},
  pages={83--104},
  year={1997},
  publisher={Annual Reviews 4139 El Camino Way, PO Box 10139, Palo Alto, CA 94303-0139, USA}
}

@inproceedings{rindt2022survival,
  title={Survival regression with proper scoring rules and monotonic neural networks},
  author={Rindt, David and Hu, Robert and Steinsaltz, David and Sejdinovic, Dino},
  booktitle={International conference on artificial intelligence and statistics},
  pages={1190--1205},
  year={2022},
  organization={PMLR}
}

@article{hochreiter1997long,
  title={Long short-term memory},
  author={Hochreiter, Sepp and Schmidhuber, J{\"u}rgen},
  journal={Neural computation},
  volume={9},
  number={8},
  pages={1735--1780},
  year={1997},
  publisher={MIT press}
}

@article{harrell1996multivariable,
  title={Multivariable prognostic models: issues in developing models, evaluating assumptions and adequacy, and measuring and reducing errors},
  author={Harrell Jr, Frank E and Lee, Kerry L and Mark, Daniel B},
  journal={Statistics in medicine},
  volume={15},
  number={4},
  pages={361--387},
  year={1996},
  publisher={Wiley Online Library}
}

@book{abergel2016limit,
  title={Limit order books},
  author={Abergel, Fr{\'e}d{\'e}ric and Anane, Marouane and Chakraborti, Anirban and Jedidi, Aymen and Toke, Ioane Muni},
  year={2016},
  publisher={Cambridge University Press}
}

@article{van2016wavenet,
  title={Wavenet: A generative model for raw audio},
  author={Van Den Oord, Aaron and Dieleman, Sander and Zen, Heiga and Simonyan, Karen and Vinyals, Oriol and Graves, Alex and Kalchbrenner, Nal and Senior, Andrew and Kavukcuoglu, Koray and others},
  journal={arXiv preprint arXiv:1609.03499},
  volume={12},
  year={2016}
}

@article{lambert2016summary,
  title={Summary measure of discrimination in survival models based on cumulative/dynamic time-dependent ROC curves},
  author={Lambert, J{\'e}r{\^o}me and Chevret, Sylvie},
  journal={Statistical methods in medical research},
  volume={25},
  number={5},
  pages={2088--2102},
  year={2016},
  publisher={SAGE Publications Sage UK: London, England}
}

@article{polsterl2020scikit,
  title={scikit-survival: A Library for Time-to-Event Analysis Built on Top of scikit-learn},
  author={P{\"o}lsterl, Sebastian},
  journal={Journal of Machine Learning Research},
  volume={21},
  number={212},
  pages={1--6},
  year={2020}
}

@article{liu2024kan,
  title={Kan: Kolmogorov-arnold networks},
  author={Liu, Ziming and Wang, Yixuan and Vaidya, Sachin and Ruehle, Fabian and Halverson, James and Solja{\v{c}}i{\'c}, Marin and Hou, Thomas Y and Tegmark, Max},
  journal={arXiv preprint arXiv:2404.19756},
  year={2024}
}

@article{lagakos1979general,
  title={General right censoring and its impact on the analysis of survival data},
  author={Lagakos, Stephen W},
  journal={Biometrics},
  pages={139--156},
  year={1979},
  publisher={JSTOR}
}

@article{vaswani2017attention,
  title={Attention is all you need},
  author={Vaswani, Ashish and Shazeer, Noam and Parmar, Niki and Uszkoreit, Jakob and Jones, Llion and Gomez, Aidan N and Kaiser, {\L}ukasz and Polosukhin, Illia},
  journal={Advances in neural information processing systems},
  volume={30},
  year={2017}
}

@article{gerds2006consistent,
  title={Consistent estimation of the expected Brier score in general survival models with right-censored event times},
  author={Gerds, Thomas A and Schumacher, Martin},
  journal={Biometrical Journal},
  volume={48},
  number={6},
  pages={1029--1040},
  year={2006},
  publisher={Wiley Online Library}
}

@article{balakrishnan2012left,
  title={Left truncated and right censored Weibull data and likelihood inference with an illustration},
  author={Balakrishnan, Narayanaswamy and Mitra, Debanjan},
  journal={Computational Statistics \& Data Analysis},
  volume={56},
  number={12},
  pages={4011--4025},
  year={2012},
  publisher={Elsevier}
}

@incollection{bouchaud2009markets,
  title={How markets slowly digest changes in supply and demand},
  author={Bouchaud, Jean-Philippe and Farmer, J Doyne and Lillo, Fabrizio},
  booktitle={Handbook of financial markets: dynamics and evolution},
  pages={57--160},
  year={2009},
  publisher={Elsevier}
}

@book{cartea2015algorithmic,
  title={Algorithmic and high-frequency trading},
  author={Cartea, {\'A}lvaro and Jaimungal, Sebastian and Penalva, Jos{\'e}},
  year={2015},
  publisher={Cambridge University Press}
}

@article{handa1996limit,
  title={Limit order trading},
  author={Handa, Puneet and Schwartz, Robert A},
  journal={The Journal of Finance},
  volume={51},
  number={5},
  pages={1835--1861},
  year={1996},
  publisher={Wiley Online Library}
}

@article{simon2011regularization,
  title={Regularization paths for Cox's proportional hazards model via coordinate descent},
  author={Simon, Noah and Friedman, Jerome H and Hastie, Trevor and Tibshirani, Rob},
  journal={Journal of statistical software},
  volume={39},
  pages={1--13},
  year={2011}
}

@article{ishwaran2008random,
  title={Random survival forests},
  author={Ishwaran, Hemant and Kogalur, Udaya B and Blackstone, Eugene H and Lauer, Michael S},
  year={2008}
}

@article{lo2002econometric,
  title={Econometric models of limit-order executions},
  author={Lo, Andrew W and MacKinlay, A Craig and Zhang, June},
  journal={Journal of Financial Economics},
  volume={65},
  number={1},
  pages={31--71},
  year={2002},
  publisher={Elsevier}
}

@inproceedings{lee2018deephit,
  title={Deephit: A deep learning approach to survival analysis with competing risks},
  author={Lee, Changhee and Zame, William and Yoon, Jinsung and Van Der Schaar, Mihaela},
  booktitle={Proceedings of the AAAI conference on artificial intelligence},
  volume={32},
  number={1},
  year={2018}
}

@article{graf1999assessment,
  title={Assessment and comparison of prognostic classification schemes for survival data},
  author={Graf, Erika and Schmoor, Claudia and Sauerbrei, Willi and Schumacher, Martin},
  journal={Statistics in medicine},
  volume={18},
  number={17-18},
  pages={2529--2545},
  year={1999},
  publisher={Wiley Online Library}
}

@article{cox1972regression,
  title={Regression models and life-tables},
  author={Cox, David R},
  journal={Journal of the Royal Statistical Society: Series B (Methodological)},
  volume={34},
  number={2},
  pages={187--202},
  year={1972},
  publisher={Wiley Online Library}
}

@book{kalbfleisch2002statistical,
  title={The statistical analysis of failure time data},
  author={Kalbfleisch, John D and Prentice, Ross L},
  year={2002},
  publisher={John Wiley \& Sons}
}

@article{ash2020warm,
  title={On warm-starting neural network training},
  author={Ash, Jordan and Adams, Ryan P},
  journal={Advances in neural information processing systems},
  volume={33},
  pages={3884--3894},
  year={2020}
}

@article{sambharya2024learning,
  title={Learning to warm-start fixed-point optimization algorithms},
  author={Sambharya, Rajiv and Hall, Georgina and Amos, Brandon and Stellato, Bartolomeo},
  journal={Journal of Machine Learning Research},
  volume={25},
  number={166},
  pages={1--46},
  year={2024}
}

@article{wang2024flexnet,
  title={FlexNet: A warm start method for deep reinforcement learning in hybrid electric vehicle energy management applications},
  author={Wang, Hanchen and Arjmandzadeh, Ziba and Ye, Yiming and Zhang, Jiangfeng and Xu, Bin},
  journal={Energy},
  volume={288},
  pages={129773},
  year={2024},
  publisher={Elsevier}
}

@article{kvamme2019time,
  title={Time-to-event prediction with neural networks and Cox regression},
  author={Kvamme, H{\aa}vard and Borgan, {\O}rnulf and Scheel, Ida},
  journal={Journal of machine learning research},
  volume={20},
  number={129},
  pages={1--30},
  year={2019}
}

@article{genet2024temporal,
  title={A temporal kolmogorov-arnold transformer for time series forecasting},
  author={Genet, Remi and Inzirillo, Hugo},
  journal={ArXiv},
  year={2024}
}

@article{genet2024tkan,
  title={Tkan: Temporal kolmogorov-arnold networks},
  author={Genet, Remi and Inzirillo, Hugo},
  journal={arXiv preprint arXiv:2405.07344},
  year={2024}
}

@article{han2024kan4tsf,
  title={Kan4tsf: Are kan and kan-based models effective for time series forecasting?},
  author={Han, Xiao and Zhang, Xinfeng and Wu, Yiling and Zhang, Zhenduo and Wu, Zhe},
  journal={arXiv preprint arXiv:2408.11306},
  year={2024}
}

@inproceedings{zhang2024transformer,
  title={Transformer-KAN: A Novel Deep Learning Model for Improving Wind Power Forecasting Accuracy by Integrating Multi-source Data},
  author={Zhang, Yajuan and Gu, Tao and Wang, Limin},
  booktitle={2024 5th International Symposium on Computer Engineering and Intelligent Communications (ISCEIC)},
  pages={49--53},
  year={2024},
  organization={IEEE}
}

@article{lundberg2017unified,
  title={A unified approach to interpreting model predictions},
  author={Lundberg, Scott M and Lee, Su-In},
  journal={Advances in Neural Information Processing Systems},
  volume={30},
  year={2017}
}

@article{kvamme2021continuous,
  title={Continuous and discrete-time survival prediction with neural networks},
  author={Kvamme, H{\aa}vard and Borgan, {\O}rnulf},
  journal={Lifetime data analysis},
  volume={27},
  number={4},
  pages={710--736},
  year={2021},
  publisher={Springer}
}

\appendix

\section{Proper Scoring Rules}\label{appendix:proper_scoring}

In Section~\ref{sec:metrics}, we noted that the negative right-censored log-likelihood (RCLL) is a 
\emph{proper scoring rule}, meaning that in expectation the true survival distribution uniquely minimizes the score \citep{rindt2022survival}. This property ensures that RCLL provides a statistically principled 
measure of model calibration. By contrast, other commonly used metrics such as the C-index and 
time-dependent AUC (discrimination), or the Brier score and its integrated form, IBS (calibration), 
are not proper scoring rules. Formally, a scoring rule $\mathcal{R}$ takes as input a predictive distribution 
$\hat{S}$ over a set $\mathcal{Y}$, with an observed sample $y \in \mathcal{Y}$, and returns a score 
$\mathcal{R}(\hat{S}, y) \in \mathbb{R}$. With positive scoring rules, higher scores indicate an 
improvement in model fit.

In survival analysis, a scoring rule $\mathcal{R}$ is called \emph{proper} if, for any true distribution $S$, 
\[
\mathbb{E}_{(T,C)|\mathbf{x}} \!\left[ \mathcal{R}(S(t \mid \mathbf{x}), (T^C, \delta)) \right] 
\;\;\geq\;\;
\mathbb{E}_{ (T,C)|\mathbf{x}} \!\left[ \mathcal{R}(\hat{S}(t \mid \mathbf{x}), (T^C, \delta)) \right],
\]
for all predictive survival function $\hat{S}$.  

Intuitively, this means that a proper scoring rule incentivizes truthful probabilistic forecasts:  
the best expected score is achieved when the predictive distribution matches the real distribution. In the context of survival analysis, \citet{rindt2022survival} show that RCLL is a proper scoring rule.  
This implies that, in expectation, the true survival distribution achieves the optimal log-likelihood.  
By contrast, widely used discrimination metrics such as the C-index or the time-dependent AUC, 
as well as calibration measures like the Integrated Brier Score (IBS), do not satisfy this property. For more details and a formal treatment of proper scoring in survival analysis, 
see \citet{rindt2022survival}.

\section{Evaluation Metrics}\label{appendix:metrics}

We provide formal definitions of the evaluation metrics used in this paper\textsuperscript{4}: AUC, C-index, and Brier Score.  
Throughout, let $\{(\mathbf{x}_i, T^C_i, \delta_i)\}_{i=1}^n$ denote the dataset, where $T^C_i = \min\{T_i, C_i\}$ is the observed time of order $i$, $\delta_i=1$ if the order is executed (partially or entirely) and $0$ if censored, and $\hat{S}(t \mid \mathbf{x}_i)$ is the model-predicted survival function.  

For evaluation, we define a \emph{time-dependent risk score} as
\[
r_i(t) = 1 - \hat{S}(t \mid \mathbf{x}_i),
\]
so that higher values correspond to higher predicted probability of execution by time $t$ \citep{lee2019dynamic}.  

\footnotetext[4]{RCLL is provided in Equation~\ref{eq:log_likelihood1}.}

\subsection{Time-dependent AUC}

The AUC at a specific horizon $t$ measures the model’s ability to distinguish between 
orders that execute before or at $t$ (cumulative cases) and those still active at $t$ (dynamic controls) 
\citep{lambert2016summary, polsterl2020scikit}.  
Formally,
\[
\mathrm{AUC}(t) =
\frac{\sum_{i=1}^n \sum_{j=1}^n 
\mathcal{I}(T^C_j > t)\, \mathcal{I}(T^C_i \le t)\, \omega_i(t)\,
\mathcal{I}\!\big(r_j(t) \le r_i(t)\big)}
{\Big(\sum_{i=1}^n \mathcal{I}(T^C_i > t)\Big)\,
 \Big(\sum_{i=1}^n \mathcal{I}(T^C_i \le t)\, \omega_i(t)\Big)} ,
\]
where $r_i(t) = 1 - \hat{S}(t \mid \mathbf{x}_i)$ is the risk score at horizon $t$, 
$\omega_i(t)$ are inverse-probability-of-censoring weights (IPCW), and 
$\mathcal{I}(\cdot)$ denotes the indicator function. This corresponds exactly to the cumulative/dynamic AUC estimator implemented in Scikit-survival \citep{polsterl2020scikit}.

Intuitively, $\text{AUC}(t)$ measures the probability that, at horizon $t$, a randomly chosen executed order receives a higher predicted risk score than a randomly chosen still-active order.

\subsection{Concordance Index (C-index)}

The concordance index ($C$) evaluates overall ranking performance across all comparable pairs \citep{uno2011c}.  
It is defined as:
\[
C = \frac{\sum_{i=1}^n \sum_{j=1}^n 
\mathcal{I}(T^C_i < T^C_j) \cdot \mathcal{I}(\delta_i = 1) \cdot 
\big[\mathcal{I}(r_i > r_j) + 0.5 \cdot \mathcal{I}(r_i = r_j)\big]}
{\sum_{i=1}^n \sum_{j=1}^n \mathcal{I}(T^C_i < T^C_j) \cdot \mathcal{I}(\delta_i = 1)} .
\]

Here $r_i$ is a fixed, order-specific risk score. When predicted risks are identical, 0.5 is counted \citep{harrell1996multivariable}. Note that the C-index expects a single risk score per order; we therefore define it as the predicted probability of execution up to the largest evaluation horizon $t^\ast$:  
\[
r_i = 1 - \hat{S}(t^\ast \mid \mathbf{x}_i).
\]

\subsection{Brier Score}

The time-dependent Brier score quantifies the squared error between predicted survival probabilities and observed outcomes at a given horizon $t$ \citep{graf1999assessment, gerds2006consistent}.  
It is defined as:
\[
\text{BS}(t) = \frac{1}{n} \sum_{i=1}^n \omega_i(t) \cdot \big(\hat{S}(t \mid \mathbf{x}_i) - \mathcal{I}(T^C_i > t)\big)^2,
\]
where $\mathcal{I}(T^C_i > t)$ indicates whether order $i$ is still active at $t$, and $\omega_i(t)$ are inverse probability of censoring weights (IPCW) \citep{polsterl2020scikit}. In experiments, we report the integrated version (IBS), obtained by averaging BS(t) across horizons.

\subsection{Proper Scoring Property}

Among the metrics above, only the right-censored log-likelihood (RCLL), 
introduced in Section~\ref{sec:metrics}, is a \emph{proper scoring rule} 
\citep{rindt2022survival}. 
This means that, in expectation, the true survival distribution minimizes the score. 
By contrast, C-index, time-dependent AUC, and the Brier Score are not proper scores 
and should be interpreted as complementary measures of discrimination or accuracy. 
For a general discussion of proper scoring rules, see \citet{rindt2022survival}.

\section{Predictive features}\label{appendix:Predictive_features}

\begin{figure}[t]
\centerline{\includegraphics[scale=0.60]{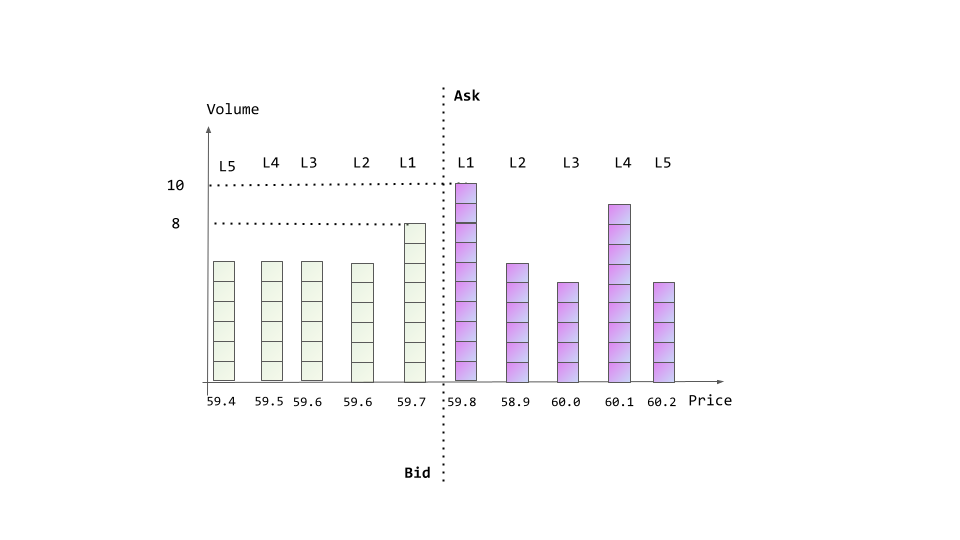}}
\caption{Example of a LOB snapshot and feature construction.}
\label{fig:example1}
\end{figure}

As we present in Section~\ref{sec:data_generation}, the input data for our model is composed of three key sources of information: {LOB Features}, {Agent Actions} and {Queue position of the order}. We now detail the three components.

 \textbf{LOB Features}: Information from the LOB including top 5 prices, cumulative volumes, and a set of derived statistics. ``Same side'' is the side of the market that the limit order is submitted to, if the limit order is to buy, the bid side of the market is the same side. The ``opposite side'' is the other side of the market that the limit order is submitted to, if the limit order is to buy, the ask side of the market is the opposite side. In addition, for the price at each level, we do not use the absolute values. Instead, we use the distance between the price of the corresponding level and the mid price: in ticks. Other predictive patterns include volatility\textsuperscript{5}, spread, volume imbalance\textsuperscript{6} and time of day. As a result, for each snapshot of a LOB we have 24 features in total to describe the snapshot of the LOB. We illustrate this in Figure~\ref{fig:example1} with a snapshot of a LOB, where the tick size is 0.1. Suppose that we are studying a limit order at the ask side, and there are 4 units before this order. As such, the same side indicates the ask side; the opposite side indicates the bid side; $mid = \frac{59.8 + 59.7}{2} = 59.75$. Then the snapshot of the LOB can be described by the vector:
\footnotetext[5]{Volatility is estimated by computing the rolling average of the squared returns of the mid-price time series over a window of 1,000 trades.}
\footnotetext[6]{Volume imbalance is computed as $VI = \frac{V_{same} - V_{opposite}}{V_{same} + V_{opposite}}$. Intuitively, a negative $VI$ indicates a stronger opposite side, indicating a higher fill probability and a faster execution of the order, vice versa.}
\[
x_m = \{p^l_s, v^l_s,p^l_o, v^l_o\}_{l=1}^n
\]

where $p^l_s, v^l_s,p^l_o, v^l_o$ denotes the price of the same side, the cumulative volume of the same side, the price of the opposite side, the cumulative volume of the opposite side for price level $l \in \{1,2,3,4,5\}$. Considering $l=1$, $p^1_s = 0.5, v^1_s = 10,p^1_o = 0.5, v^1_o = 8$; considering $l=2$, $p^2_s = 1.5, v^2_s = 10+6 = 16,p^2_o = 1.5, v^2_o = 8 + 6 = 14$, etc.

\textbf{Agent Actions}: The actions of other agents in the market associated with each evolution of the LOB. The action types include: 
\begin{itemize}
                    \item \( I \): insertion of a new limit order,
                    \item \( C \): cancellation of an outstanding limit order,
                    \item \( R \): modification of a limit order that loses priority,
                    \item \( r \): modification of a limit order that keeps priority,
                    \item \( S \): modification of a limit order that becomes aggressive,
                    \item \( T \): an aggressive order (market or limit) that is immediately executed,
                    \item \( J \): insertion of a stop order.
                \end{itemize}

These actions provide contextual insights into the market dynamics. Moreover, we have access to the CAC 40 data from Euronext, we derive 5 representative statistics of the submitting agent, which include Limit Ratio, Market Ratio, Cancel Ratio, Trade Ratio, Aggressive Trade Ratio.

 \textbf{Queue position of the order}: The queue position of the order being studied. The LOB features and agent actions can be considered the market context, we also add the queue position of the order being studied. The motivation is that queue position of the order influences its execution \citep{huang2015simulating, cartea2014buy}. Each order is associated with its $q$, which is used solely in the predictor in Figure~\ref{fig:model}. For example, for an order being studied, and there are 4 units before this order to be executed, then the $queue = 4$.

\section{Hyperparameters}\label{appendix:hyperparameters}

 We perform a grid search of hyperparameters over the following ranges:  
\begin{itemize}
    \item {Regularization (weight decay):} $\{10^{-3}, 10^{-4}, 10^{-5}\}$  
    \item {Batch size:} $\{256, 512, 1024\}$
    \item {Attention heads:} \{2, 4\}  
    \item {Hidden size:} \{8, 16, 32\}  
    \item {KAN grid size:} \{5, 7\}  
    \item {Spline order:} \{3\}  
    \item {Number of layers:} \{2, 4, 6\}  
    \item {Dropout rate:} \{0.1, 0.2, 0.3, 0.4, 0.5\}  
    \item {Action type embedding size:} \{2, 4, 8, 16\}  
\end{itemize}  
For the Dilated Causal Convolutions (DCC) applied to LOB snapshots, we fix the kernel size to 3 and the dilation factor to 1, following \citet{arroyo2024deep}.

\section{Results of larger prediction horizon}\label{appendix:long}

In Section~\ref{sec:experiments}, we presented the average values of evaluation metrics computed over a set of 20 prediction time points, selected between the 10th and 50th percentiles (approximately 0.627 seconds) of the event time distribution.

\begin{figure}[t]
    \centering
    \includegraphics[width=1\textwidth]{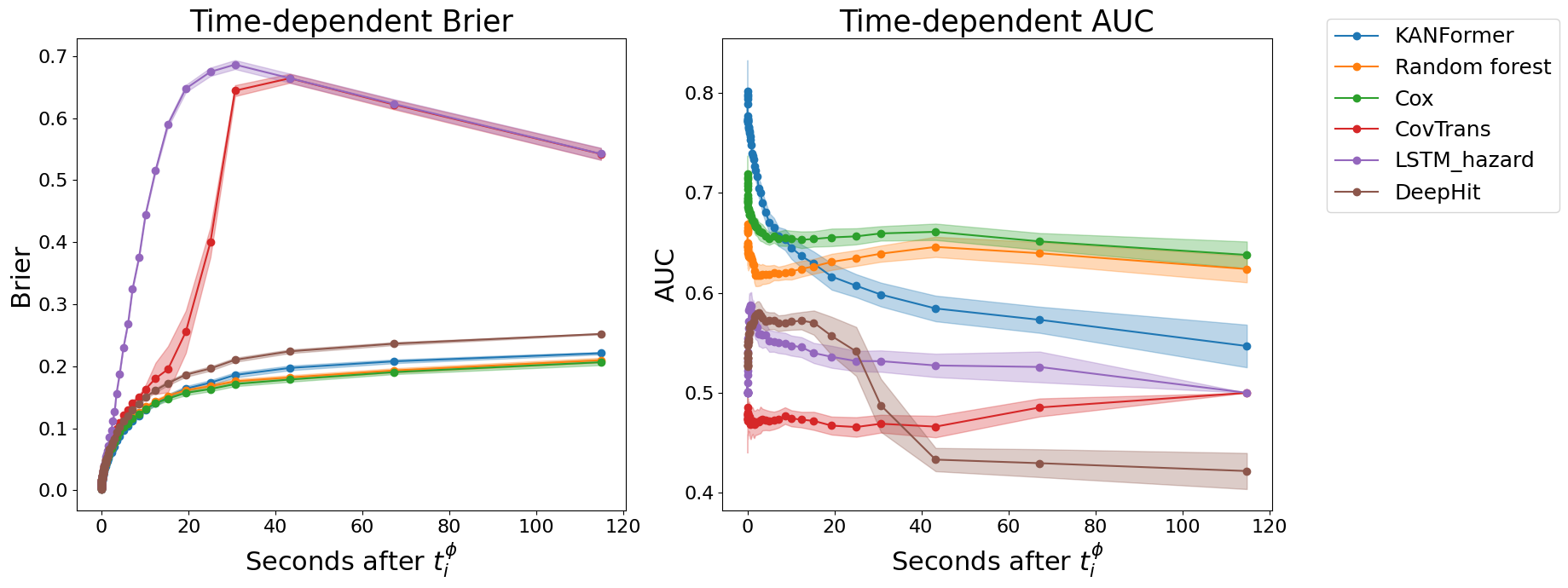}
    \caption{Time-dependent Brier score and AUC over larger prediction horizons.}
    \label{Fig.Evolution_metrics3}
\end{figure}

As a complementary analysis, Figure~\ref{Fig.Evolution_metrics3} shows the evolution of BS, AUC for longer prediction horizons. We observe that as the prediction horizon increases, the performance of most models declines. In particular, AUC values approach 0.5, indicating near-random discrimination. This result is expected: in a high-frequency trading context, a prediction horizon of 120 seconds is relatively long, making it difficult for models to accurately predict survival probabilities and density functions over such extended timeframes. An additional noteworthy finding is that the two non-deep learning baselines, Random Forest and the Cox model, perform better than deep learning models at longer horizons, suggesting that traditional models may offer greater robustness for long-range predictions.

Importantly, our model, KANFormer, achieves higher discrimination performance at short prediction horizons, making it well-suited to high-frequency contexts where timely and precise predictions are critical. This insight suggests applying our model iteratively with short prediction intervals. This strategy ensures that the model operates within the regime where its discrimination ability remains strong, thereby enabling reliable, real-time fill probability estimation.

\end{document}